%% file: main.tex
\documentclass[letterpaper, 10 pt, conference]{ieeeconf}
\IEEEoverridecommandlockouts
\input{preamble}

\title{\LARGE \bf
LiDAR Data Synthesis with Denoising Diffusion Probabilistic Models
}

\author{Kazuto Nakashima$^{1}$ and Ryo Kurazume$^{1}$
\thanks{*This work was supported by JSPS KAKENHI Grant Number JP23K16974 and JST [Moonshot R\&D] [Grant Number JPMJMS2032]}
\thanks{$^{1}$Kazuto Nakashima and Ryo Kurazume are with the Faculty of Information Science and Electrical Engineering, Kyushu University, Japan. {\tt\small \{k\_nakashima,kurazume\}@ait.kyushu-u.ac.jp}}}

\begin{document}

\maketitle
\thispagestyle{empty}
\pagestyle{empty}

\begin{abstract}
	Generative modeling of 3D LiDAR data is an emerging task with promising applications for autonomous mobile robots, such as scalable simulation, scene manipulation, and sparse-to-dense completion of LiDAR point clouds.
	While existing approaches have demonstrated the feasibility of image-based LiDAR data generation using deep generative models, they still struggle with fidelity and training stability.
	In this work, we present R2DM, a novel generative model for LiDAR data that can generate diverse and high-fidelity 3D scene point clouds based on the image representation of range and reflectance intensity.
	Our method is built upon denoising diffusion probabilistic models (DDPMs), which have shown impressive results among generative model frameworks in recent years.
	To effectively train DDPMs in the LiDAR domain, we first conduct an in-depth analysis of data representation, loss functions, and spatial inductive biases.
	Leveraging our R2DM model, we also introduce a flexible LiDAR completion pipeline based on the powerful capabilities of DDPMs.
	We demonstrate that our method surpasses existing methods in generating tasks on the KITTI-360 and KITTI-Raw datasets, as well as in the completion task on the KITTI-360 dataset.
	Our project page can be found at \url{https://kazuto1011.github.io/r2dm}.
\end{abstract}

\section{Introduction}
\label{sec:introduction}

LiDAR (light detection and ranging) sensors play a pivotal role in enabling mobile robots to perceive surrounding obstacles and geometry for autonomous navigation.
The sensors produce accurate point clouds of 3D scenes by emitting laser beams at omnidirectional angles and detecting reflections from objects.
The collected point clouds, along with laser reflectance information, can be utilized for 3D scene understanding tasks~\cite{li2020deep}, such as semantic/instance segmentation and object detection.

However, high-density and high-quality point clouds are not readily accessible on all platforms because increasing the number of beams elevates cost and energy consumption.
Therefore, using low-cost yet low-beam LiDAR sensors can be a practical option for system development, while they lead to performance degradation in 3D scene understanding due to the sparsity of the point clouds.

Generative modeling of LiDAR point clouds is a frontier approach to tackle these issues, which aims to learn the prior distribution of 3D scenes.
The sparse or incomplete point clouds can be restored by using the learned priors.
Motivated by the impressive achievements of deep generative models~\cite{bond2022deep}, various approaches have been proposed for LiDAR data generation~\cite{caccia2019deep,nakashima2021learning,nakashima2023generative,zyrianov2022learning} using variational autoencoders (VAEs)~\cite{kingma2014auto}, generative adversarial networks (GANs)~\cite{goodfellow2014generative}, and diffusion models~\cite{song2019generative,song2020improved,song2021scorebased,ho2020denoising,nichol2021improved,kingma2021variational,saharia2022photorealistic}.

In this paper, we present R2DM, a novel generative model for LiDAR data that can generate diverse and high-fidelity 3D scene point clouds.
As a framework for building a generative model, we employ denoising diffusion probabilistic models (DDPMs)~\cite{ho2020denoising}, one of the most successful approaches among diffusion models.
DDPMs offer various benefits over other frameworks, such as training stability, sample quality, and versatility to inverse problems, as demonstrated in the wide range of domains.
We demonstrate the effectiveness of DDPMs on generative modeling of LiDAR data.

\begin{figure}[t]
	\centering
	\footnotesize
	{\renewcommand{\arraystretch}{3}%
		\begin{tabularx}{\hsize}{CC}
			\includegraphics[align=c,width=\hsize]{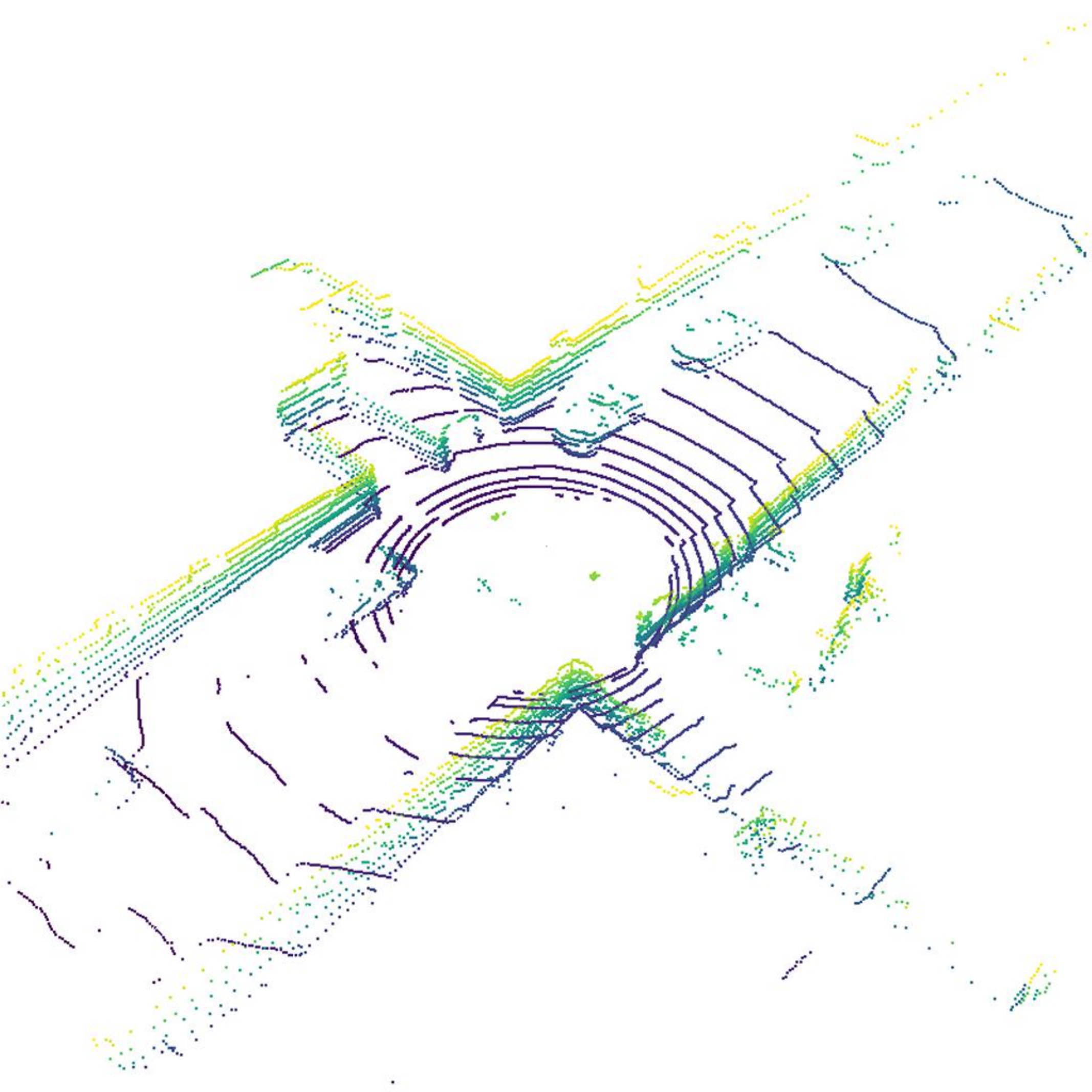}   & \includegraphics[align=c,width=\hsize]{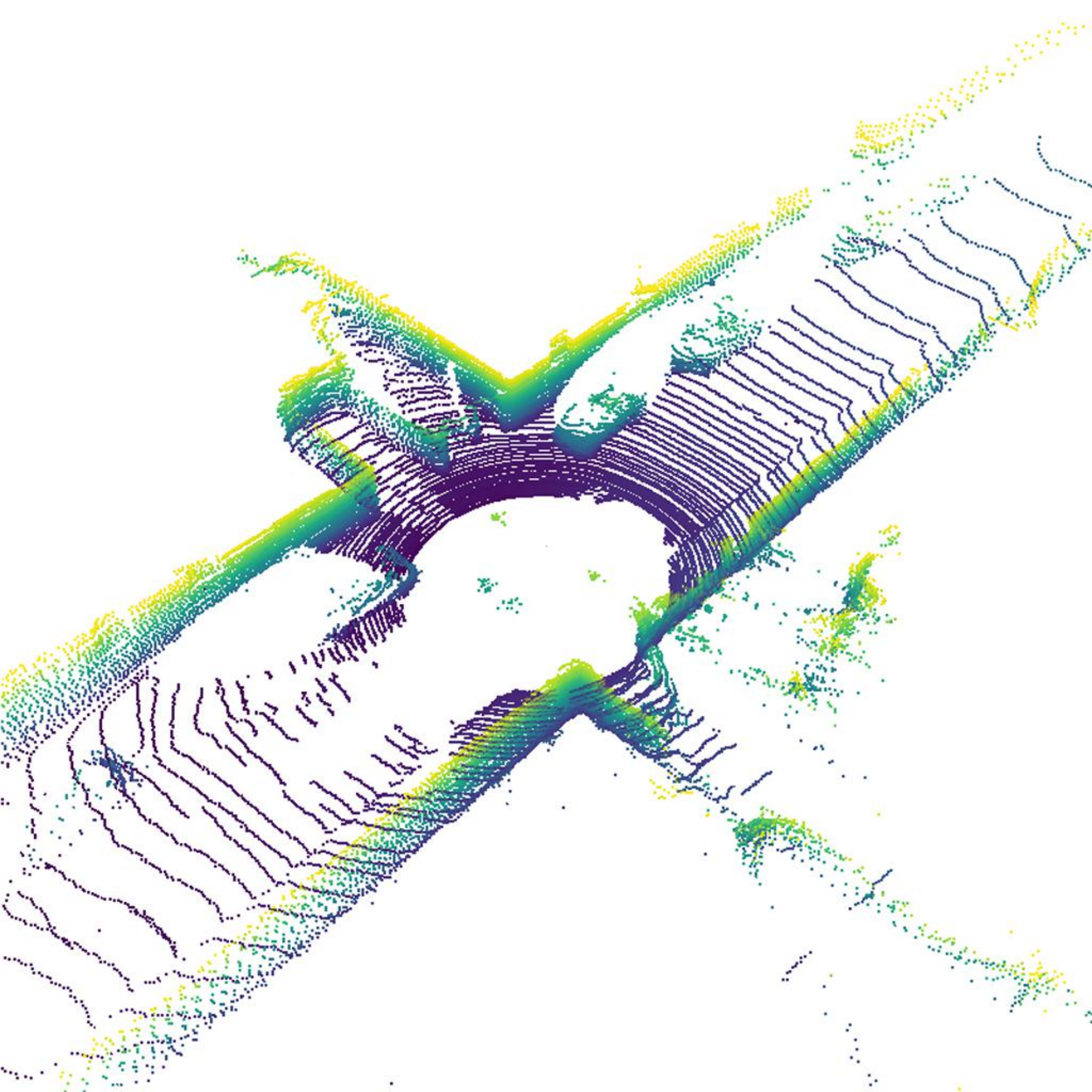}   \\
			\includegraphics[align=c,width=\hsize]{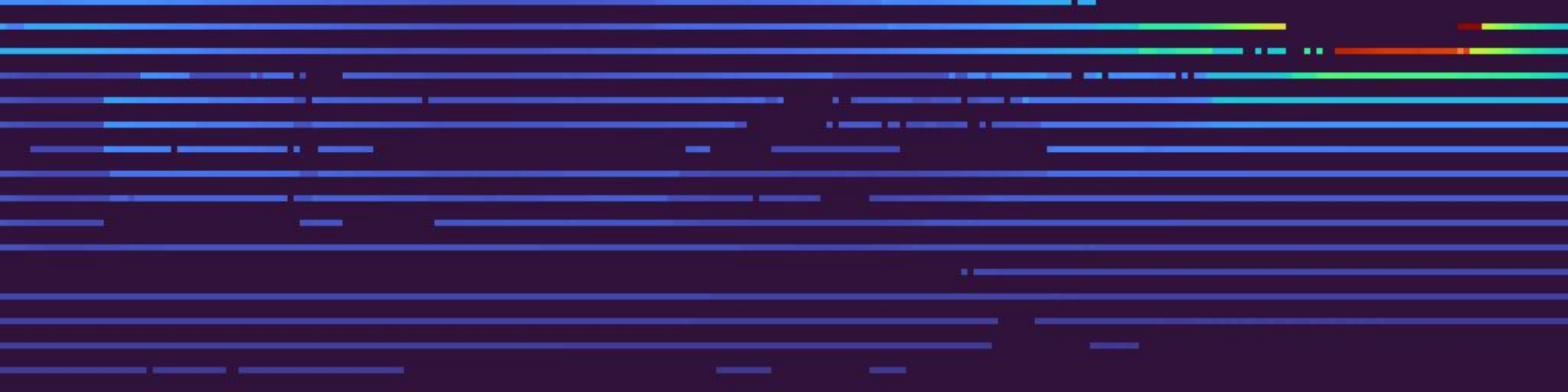} & \includegraphics[align=c,width=\hsize]{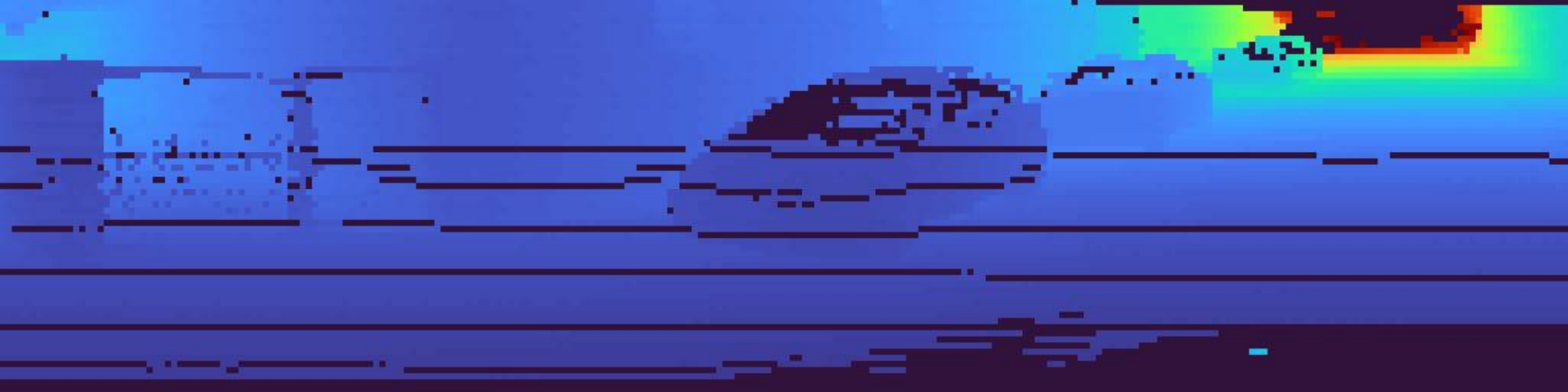} \\
			\includegraphics[align=c,width=\hsize]{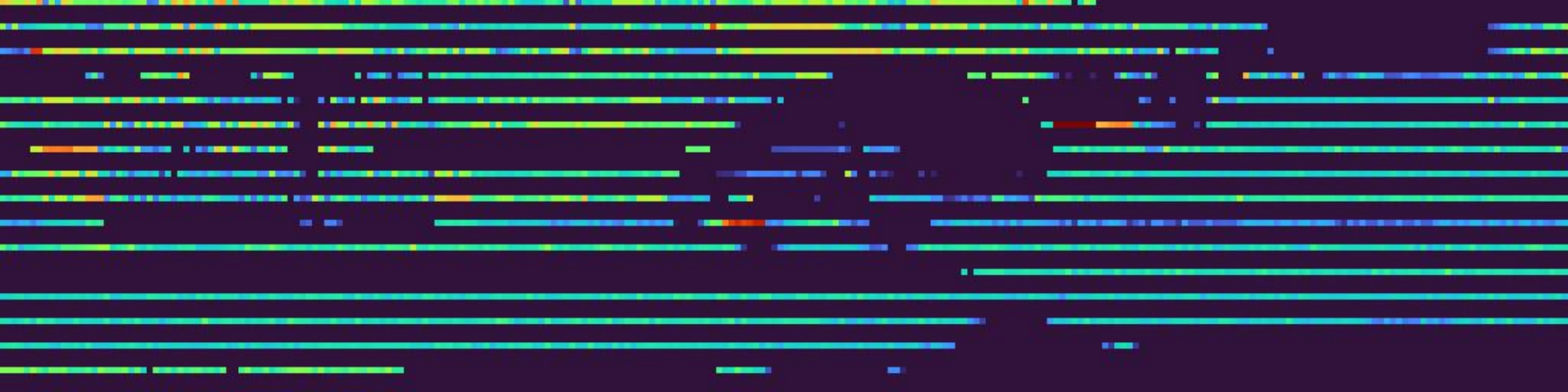} & \includegraphics[align=c,width=\hsize]{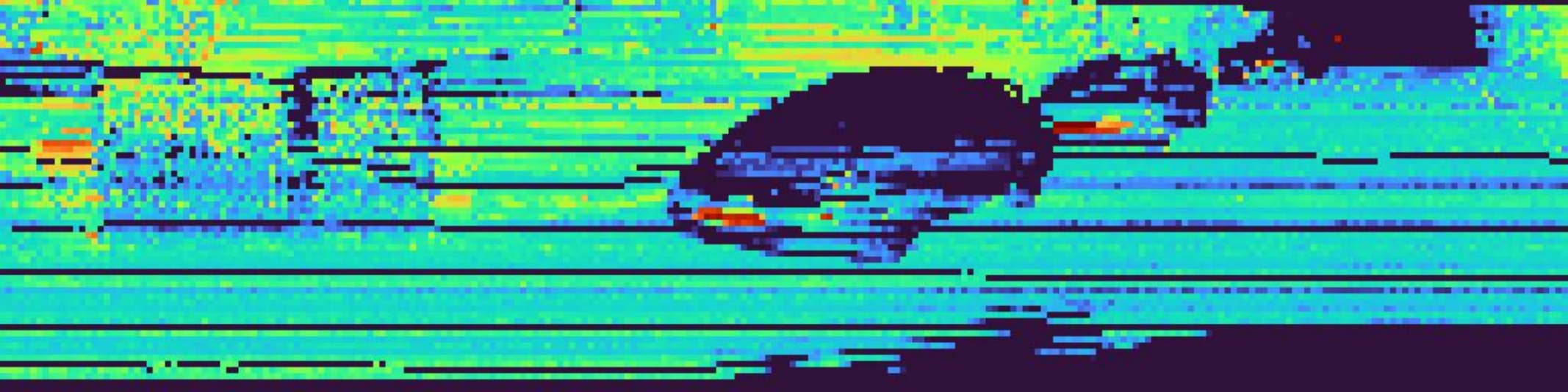} \\
		\end{tabularx}}
	\\ \vspace{1mm}
	\begin{tabularx}{\hsize}{CC}
		Sparse inputs (16 beams) & Our results (64 beams) 
	\end{tabularx}
	\caption{\textbf{LiDAR upsampling using our diffusion model}. We show the sparse LiDAR inputs and our upsampled results as point clouds (top), range images (middle), and reflectance images (bottom). Our results were obtained through image-based conditional generation using our model. The range and reflectance images are partially zoomed for visual purposes.}
	\label{fig:first_figure}
\end{figure}

Similar to the relevant studies~\cite{caccia2019deep,nakashima2021learning,nakashima2023generative,zyrianov2022learning}, we cast our task as an image-based generation, wherein LiDAR point clouds are represented by equirectangular images where each pixel contains pointwise range and the corresponding laser reflectance values.
To gain the performance of DDPMs on LiDAR data, we first investigate the suitable model design in three aspects: loss functions, data representation, and spatial inductive bias.
With our designed architecture, R2DM achieves state-of-the-art generation performance across various levels of metrics, including point clouds, range images, and bird's eye views.
Furthermore, the pre-trained R2DM can be applied to LiDAR completion tasks.
We propose a flexible pipeline capable of handling various types of corruption, including issues stemming from low-beam LiDAR configurations (see Fig.~\ref{fig:first_figure} for instance).

Our contributions can be summarized as follows:

\begin{itemize}
	\item We present R2DM (range–reflectance diffusion model), a DDPM-based generative model capable of generating diverse and high-fidelity LiDAR range images with a reflectance modality, which demonstrates state-of-the-art performances on the KITTI-360~\cite{liao2022kitti} and KITTI-Raw~\cite{geiger2013vision} datasets.
	\item We provide an in-depth analysis of the effective training of DDPMs in the LiDAR domain. Our key finding is that an explicit spatial bias significantly influences the fidelity of generated samples.
	\item We also introduce a flexible LiDAR completion pipeline using powerful properties offered by DDPMs, which also outperforms baseline methods on the beam-level upsampling task using KITTI-360.
\end{itemize}

\section{Related Work}
\label{sec:related_work}

\subsection{Generative models}

Generative models aim to learn the underlying distribution of a dataset.
During the last decade, the rapid advancements in deep neural networks have spurred the development of various frameworks for generative models.

One of the most prominent frameworks is generative adversarial networks (GANs)~\cite{goodfellow2014generative}.
A GAN consists of a pair of competing neural networks: a generator and a discriminator, which are trained alternately to minimize the adversarial objective.
While GANs are known for their sampling efficiency and high synthesis quality, training instability poses a challenge due to their competitive nature.

Recently, diffusion models, including score-based generative models~\cite{song2019generative,song2020improved,song2021scorebased} and denoising diffusion probabilistic models (DDPM)~\cite{ho2020denoising,saharia2022photorealistic,kingma2021variational,nichol2021improved}, have garnered substantial attention, especially in the field of text-to-image generation~\cite{saharia2022photorealistic}.
Diffusion models have notable advantages that have led to significant advancements.
Unlike GANs, they offer stable training with a simple objective function thanks to the formulation approximating likelihood maximization.
Training diffusion models is also efficient and scalable since a single neural network suffices to formulate both generative and inference processes.
Moreover, the generative process is learned as \textit{iterative refinement}; the task imposed on the model gets easier.
Finally, it is worth noting that pre-trained models can be adapted to solving inverse problems, such as colorization, super-resolution, and inpainting~\cite{lugmayr2022repaint}.

\subsection{LiDAR data generation}

Motivated by the progress in the natural image domain, some studies have tackled generative modeling of LiDAR data based on the \textit{range image}-based representation.
Caccia~\etal~\cite{caccia2019deep} initiated pivotal work on this subject, utilizing VAEs~\cite{kingma2014auto} and GANs~\cite{goodfellow2014generative} to train LiDAR range images.
Several studies focused on the discrete dropout noise spread on the range images, called raydrop.
Nakashima and Kurazume~\cite{nakashima2021learning} introduced DUSty, a noise-robust GAN architecture disentangling the noisy range images into denoised image space and the corresponding dropout probability.
Building upon the idea of DUSty, Nakashima~\etal~\cite{nakashima2023generative} introduced DUSty v2, a neural field-based architecture capable of representing range images at arbitrary resolutions.
They also introduced a sim2real application using the learned priors.
Despite the steady improvements, these studies~\cite{caccia2019deep,nakashima2021learning,nakashima2023generative} were limited to generating the range modality alone.

LiDARGen proposed by Zyrianov~\etal~\cite{zyrianov2022learning} is closely related to our work, which is a diffusion model that trains both range and reflectance modalities based on the image representation.
In particular, they employed NCSNv2~\cite{song2020improved}, one of the score-based generative models.
NCSNv2 expresses the data distribution $p(\bm{x})$ by training the gradient of log probability $\nabla_{\bm{x}}\log{p(\bm{x})}$ involving multi-level noise perturbation, the so-called \textit{score}.
Sampling data is performed by Langevin dynamics at stepping-down noise levels.
They also proposed to modify the model architecture, incorporating circulating kernel operations and introducing a spatial bias.
LiDARGen has demonstrated state-of-the-art results on standard LiDAR datasets, albeit with only subtle improvements compared to existing approaches.
As addressed in~\cite{zyrianov2022learning}, low sampling efficiency also remains an issue, which is dominated by the number of sampling timesteps determined when training and an evaluation cost per step using the neural network.

The aim of this paper is to improve both the fidelity and efficiency of the diffusion-based approach.
Our method R2DM is built upon the DDPM framework with timestep-agnostic training and an efficient neural network architecture.
Our experiments reveal that R2DM outperforms LiDARGen with fewer sampling steps and faster network evaluation.

\section{Proposed Method}
\label{sec:proposed_method}

To explore the effective DDPM design for LiDAR data, this section provides the formulation of DDPMs and introduce some modification regarding loss function, data representation, and spatial inductive bias.
Lastly, we also introduce the LiDAR completion pipeline using our model.

\subsection{Preliminary}

In this paper, we employ the DDPM framework that formulates transitions between data and latent spaces in continuous time $t\in[0, 1]$~\cite{kingma2021variational,saharia2022photorealistic,emiel2023simple}.
Unlike the discrete-time diffusion models~\cite{song2019generative,song2020improved,ho2020denoising} including LiDARGen~\cite{zyrianov2022learning}, there is no need to carefully define the number of steps in the training phase, which can be determined in the sampling phase afterward, balancing computational tradeoffs.

\begin{figure*}[t]
	\centering
	\footnotesize
	\begin{tabularx}{\hsize}{CC}
		\includegraphics[height=80px]{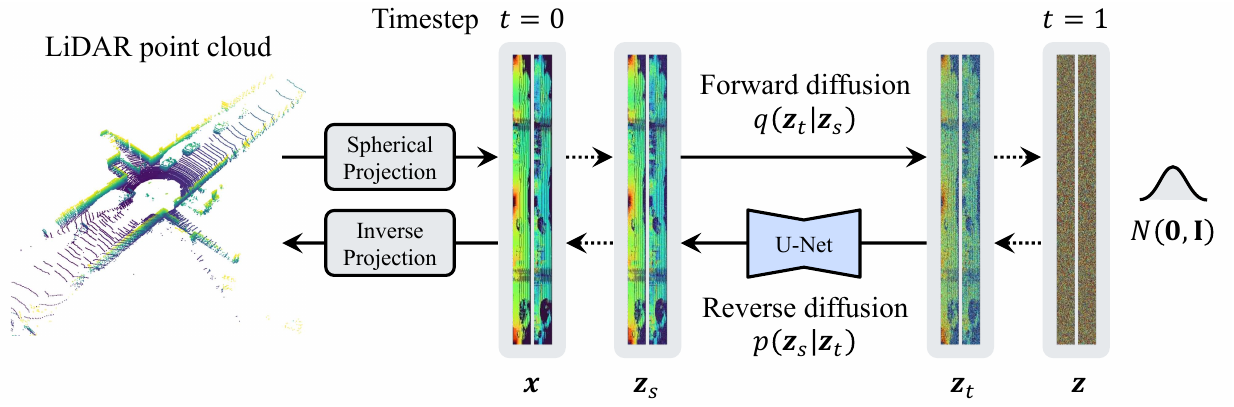} & \includegraphics[height=80px]{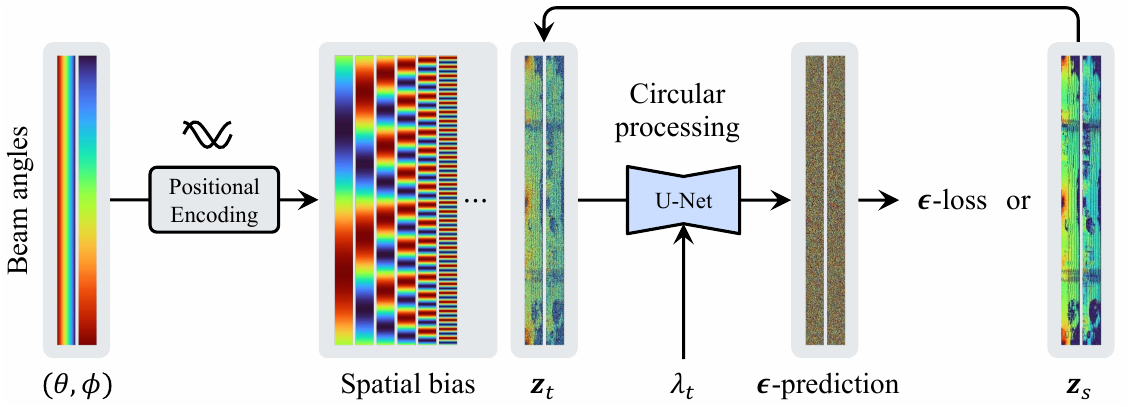} \\
		(a) Range/reflectance image-based diffusion model         & (b) Details of the reverse diffusion process          
	\end{tabularx}
	\caption{\textbf{Overview of R2DM}. (a) The diffusion processes are performed on the range/reflectance image representation. (b) U-Net is trained to recursively denoise the latent variables $z_t$ at $t>0$, conditioned by the beam angle-based spatial bias and the scheduled signal-to-noise ratio $\lambda_t$.}
	\label{fig:model}
\end{figure*}

In DDPMs, an inference process is called a forward diffusion process, which gradually corrupts the data sample $\bm{x}$ by adding Gaussian noise as $t$ progresses from 0 to 1, thereby resulting in the latent variable $\bm{z}\sim\mathcal{N}(\mathbf{0},\mathbf{I})$.
In contrast, a generative process is called a \textit{reverse} diffusion process, which gradually denoise the latent variable $\bm{z}$ at $t=1$ to be the data sample $\bm{x}$ at $t=0$.
The transitional noisy samples $\bm{z}_t$ for $0<t<1$ are also considered as the latent variables.
The schematic diagram is depicted in Fig.~\ref{fig:model}(a).

\subsubsection{Forward diffusion process}

Conveniently, since the forward diffusion process follows the additive Gaussian, the noisy samples $\bm{z}_t$ at arbitrary timestep $t$ can be given by:
\begin{equation}
	\label{eq:q_x_to_zt}
	q(\bm{z}_t \mid \bm{x})=\mathcal{N}(\alpha_t \bm{x}, \sigma_t^2 \mathbf{I}),
\end{equation}
where $\alpha_t$ and $\sigma_t$ are parameters to determine the noising schedule.
For example, the most popular schedule is $\alpha$-cosine schedule~\cite{nichol2021improved} where $\alpha_t=\mathrm{cos}(\pi t / 2)$ and $\sigma_t=\mathrm{sin}(\pi t / 2)$.
This transition distribution can be re-parameterized as:
\begin{equation}
	\label{eq:q_x_to_zt_reparam}
	\bm{z}_t=\alpha_t \bm{x} + \sigma_t \bm{\epsilon},
\end{equation}
where $\bm{\epsilon}\sim\mathcal{N}(\mathbf{0},\mathbf{I})$ and the signal-to-noise ratio of $\bm{z}_t$ can be defined as $\lambda_{t}=\alpha_t^2/\sigma_t^2=\mathrm{cot}^2(\pi t/2)$.
In addition, the transition of latent variables $q(\bm{z}_t\mid\bm{z}_s)$ from timestep $s$ to $t$, for any $0\leq s < t\leq1$, can be written as:
\begin{equation}
	\label{eq:q_zs_to_zt}
	q(\bm{z}_t\mid\bm{z}_s)=\mathcal{N}(\alpha_{t \mid s} \bm{z}_s, \sigma_{t \mid s}^2 \mathbf{I}),
\end{equation}
where $\alpha_{t \mid s}=\alpha_t/\alpha_s$ and $\sigma_{t \mid s}^2=\sigma_t-\alpha_{t \mid s}\sigma_{s}$.

\subsubsection{Reverse diffusion process}

Given the distributions above, the reverse diffusion process $p(\bm{z}_s \mid \bm{z}_t)$ is given by:
\begin{align}
	\begin{split}                                                                                                                                
	\label{eq:p_zt_to_zs}                                                                                                                        
	p(\bm{z}_s \mid \bm{z}_t) = \mathcal{N}(\bm{\mu}_t(\bm{x},\bm{z}_t),\,\Sigma_t^2 \mathbf{I}),                                                \\
	\bm{\mu}_t(\bm{x},\bm{z}_t)=\frac{\alpha_{t\mid s} \sigma_s^2}{\sigma_t^2} \bm{z}_t + \frac{\alpha_s \sigma_{t\mid s}^2}{\sigma_t^2} \bm{x}, 
	\qquad \Sigma_t^2=\frac{\sigma_{t\mid s}^2 \sigma_s^2}{\sigma_t^2}.                                                                          
	\end{split}                                                                                                                                  
\end{align}

\subsubsection{Training}

The training objective of DDPM is to estimate the unknown $\bm{x}$ in Eq.~\ref{eq:p_zt_to_zs} by a neural network, where U-Net~\cite{ronneberger2015unet} is generally used.
In general, $\bm{\epsilon}$-prediction and $\bm{\epsilon}$-loss~\cite{ho2020denoising} are preferable; re-parameterizing $\bm{x}$ as a function of noise $\bm{\epsilon}$ by Eq.~\ref{eq:q_x_to_zt_reparam}. The loss function is given by:
\begin{equation}
	\label{eq:loss}
	\mathcal{L} = \mathbb{E}_{\bm{x},\bm{\epsilon}\sim\mathcal{N}(\mathbf{0},\mathbf{I}),t\sim\mathcal{U}(0, 1)}\left[\| \bm{\epsilon} - \hat{\bm{\epsilon}}(\bm{z}_t,\lambda_{t}) \|^2_2\right],
\end{equation}
where $\hat{\bm{\epsilon}}(\cdot)$ is the neural network predicting the noise $\bm{\epsilon}$ from $\bm{z}_t$ and the corresponding $\lambda_{t}$.

\subsubsection{Sampling}

Once the training is complete, we can sample data by recursively evaluating $p({\bm{z}_s}\mid{\bm{z}_t})$ where $\bm{x}$ is approximated by $\hat{\bm{x}}=(\bm{z}_t-\sigma_t \hat{\bm{\epsilon}}(\bm{z}_t,\lambda_{t}))/\alpha_t$ with a finite number of steps $T$ from $t=1$ to $t=0$.

\subsection{Loss function}
\label{sec:loss_function}

In Eq.~(\ref{eq:loss}), the $L_2$ loss is used to supervise the model prediction. Meanwhile, in the context of monocular depth estimation, Saxena~\etal~\cite{saxena2023monocular} found that an $L_1$-based formulation offers better performance due to its robustness to large depth values and noisy holes. In our experiment, we also evaluate the $L_1$ loss and the combination of $L_1$ and $L_2$, the Huber loss.

\subsection{Data representation}
\label{sec:data_representation}

We assume a LiDAR sensor that has an angular resolution of $W$ in azimuth and $H$ in elevation and measures the range and reflectance at each angle $(\theta, \phi)$.
Then, $HW$ sets of the range and reflectance values can be projected to an equirectangular image space $\bm{x}\in\mathbb{R}^{2\times{H}\times{W}}$ by spherical projection.
Here we focus on the range representation and compare three approaches.
Similar to LiDARGen~\cite{zyrianov2022learning}, we first convert the range $\bm{d}\in[0, d_{\rm{max}}]^{{H}\times{W}}$ into the log-scale representation $\bm{d}_{\rm{log}}\in[0, 1]^{{H}\times{W}}$  as follows:
\begin{equation}
	\bm{d}_{\rm{log}}=\frac{\mathrm{log}(\bm{d}+1)}{\mathrm{log}(d_{\mathrm{max}}+1)},
\end{equation}
where $d_{\rm{max}}$ is the maximum range value.
$\bm{d}_{\rm{log}}$ is then re-scaled linearly into $[-1,1]$ as conventional DDPMs~\cite{ho2020denoising}.
This log-scale representation gains the resolution of nearby points filling a large portion of the image.
We also compare other popular representations of \textit{depth} maps: standard metric depth $\bm{d}$ and inverse depth (reciprocal)~\cite{nakashima2021learning,nakashima2023generative}, which are also normalized into $[-1, 1]$.

\subsection{Spatial inductive bias}
\label{sec:spatial_inductive_bias}

The generated pixels must be accurate and well-aligned with the angular coordinates $\{(\theta,\phi)\}$ to ensure the geometric fidelity in transformed point clouds.
Since the initial input $\bm{z}$ of the neural network $\hat{\bm{\epsilon}}(\cdot)$ is \textit{i.i.d.} Gaussian noise, an implicit positional encoding by zero padding~\cite{xu2021positional,choi2021toward} is considered a mainstay for obtaining structured outputs.
LiDARGen~\cite{zyrianov2022learning} proposed concatenating the raw angular coordinates with the input as an explicit spatial inductive bias.
However, we hypothesize that the coordinates alone, lacking in high-frequency components and horizontal continuity, are insufficient for representing detailed structures of LiDAR data.

We here generalize the spatial bias as a positional encoding, which can be seen as an identity function in the LiDARGen case.
We further investigate \textit{wideband} and \textit{cylindrical} mappings: spherical harmonics~\cite{verbin2022ref, zhang2022curl} and Fourier features~\cite{tancik2020fourier}.
Spherical harmonics represent functions on a sphere through a series of orthogonal functions, which is demonstrated in novel view synthesis~\cite{verbin2022ref} and LiDAR compression~\cite{zhang2022curl}.
We encode the angles by spherical harmonic basis functions.
We also compare Fourier features~\cite{tancik2020fourier}.
We employ a $\rm{log}_2$-spaced scheme~\cite{mildenhall2020nerf}, which extends the elevation and azimuth angles separately to a set of power-of-two frequencies so that the mapping preserves the horizontal continuity.
Fig.~\ref{fig:model}(b) depicts the mechanism.

\subsection{Noise prediction model}
\label{sec:noise_prediction_model}

Following the standard diffusion models~\cite{song2019generative,song2020improved,ho2020denoising,nichol2021improved,song2021scorebased,kingma2021variational,saharia2022photorealistic}, we use a U-Net architecture~\cite{ronneberger2015unet} to predict the noise $\bm{\epsilon}$ in Eq.~\ref{eq:loss}.
Specifically, our model is built upon Efficient U-Net~\cite{saharia2022photorealistic} which was demonstrated on monocular depth estimation~\cite{saxena2023monocular}.
We change the input and output channels as to process the 2-channel LiDAR imagery: range and reflectance.
A logarithmic signal-to-noise ratio $\log \lambda_t$ is embedded into affine weights of every adaptive group normalization (AdaGN)~\cite{dhariwal2021diffusion}.
We adopt three residual blocks at each resolution and the multi-head self-attention layer at the lowest resolution.
The final model involves 31.1M parameters which was adjusted to roughly align with LiDARGen~\cite{zyrianov2022learning} with 29.7M parameters for fair comparison.
Table~\ref{tab:model_comparison} shows the architecture comparison.
All kernel operations, such as convolution and up/down-resampling, are modified to use circular padding in the horizontal direction, as demonstrated in LiDAR processing tasks~\cite{nakashima2018learning,nakashima2021learning,nakashima2023generative,zyrianov2022learning,schubert2019circular}.

\begin{table}[t]
	\centering
	\begin{threeparttable}
		\caption{Architecture comparison of diffusion-based models}
		\label{tab:model_comparison}
		\begin{tabularx}{\hsize}{Llcc}
			\toprule
			Method                               & U-Net architecture                                           & \# params  & msec/step$^\dag$ \\
			\midrule
			LiDARGen~\cite{zyrianov2022learning} & RefineNet~\cite{lin2017refinenet}~in~\cite{song2020improved} & 29,694,082 & 47.17            \\
			\textbf{R2DM (ours)}                 & Efficient U-Net~\cite{saharia2022photorealistic}             & 31,099,650 & \bf 15.77        \\
			\bottomrule
		\end{tabularx}
		\begin{tablenotes}
			\item \dag~Average time of 1000 runs on our GPU w/ PyTorch compilation.
		\end{tablenotes}
	\end{threeparttable}
\end{table}

\subsection{LiDAR completion}
\label{sec:lidar_completion}

\begin{figure}[t]
	\centering
	\footnotesize
	\begin{tabularx}{\hsize}{rCCC}
		       & 8/64 beams                                                                 & 90\% dropout                                                                    & road removal                                                                   \\
		\cmidrule(lr){2-2} \cmidrule(lr){3-3} \cmidrule(lr){4-4}
		Input  & \includegraphics[align=c,width=\hsize]{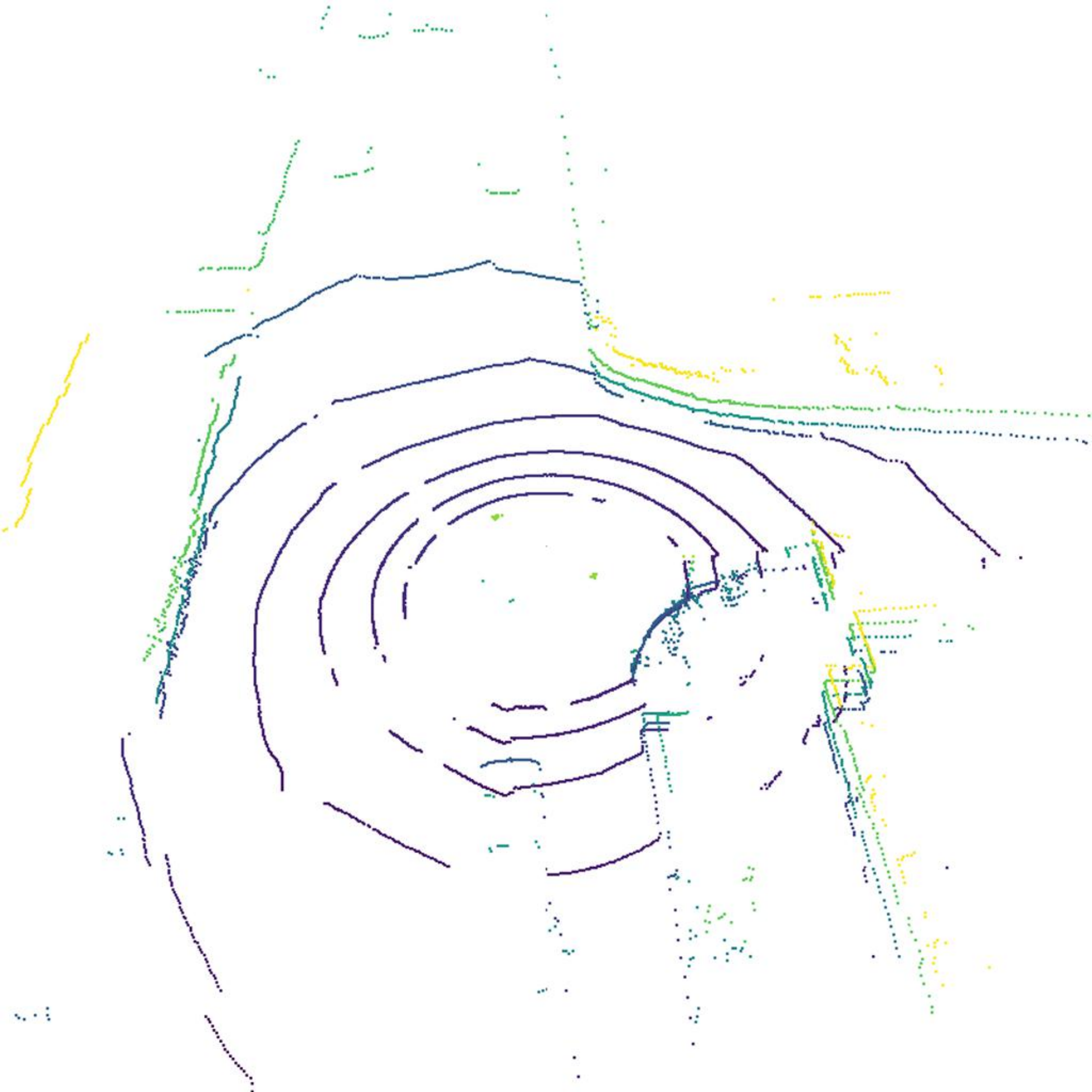}  & \includegraphics[align=c,width=\hsize]{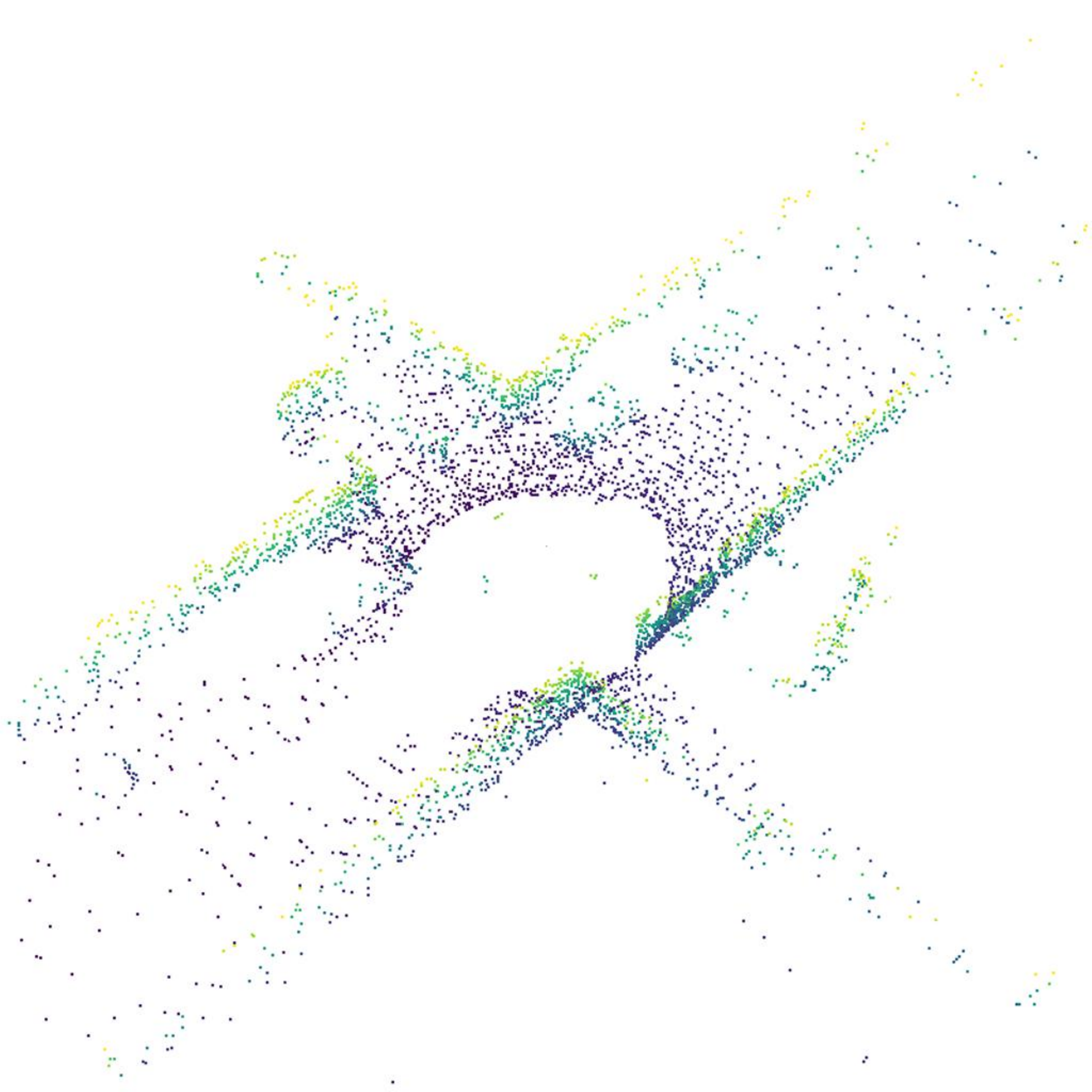}  & \includegraphics[align=c,width=\hsize]{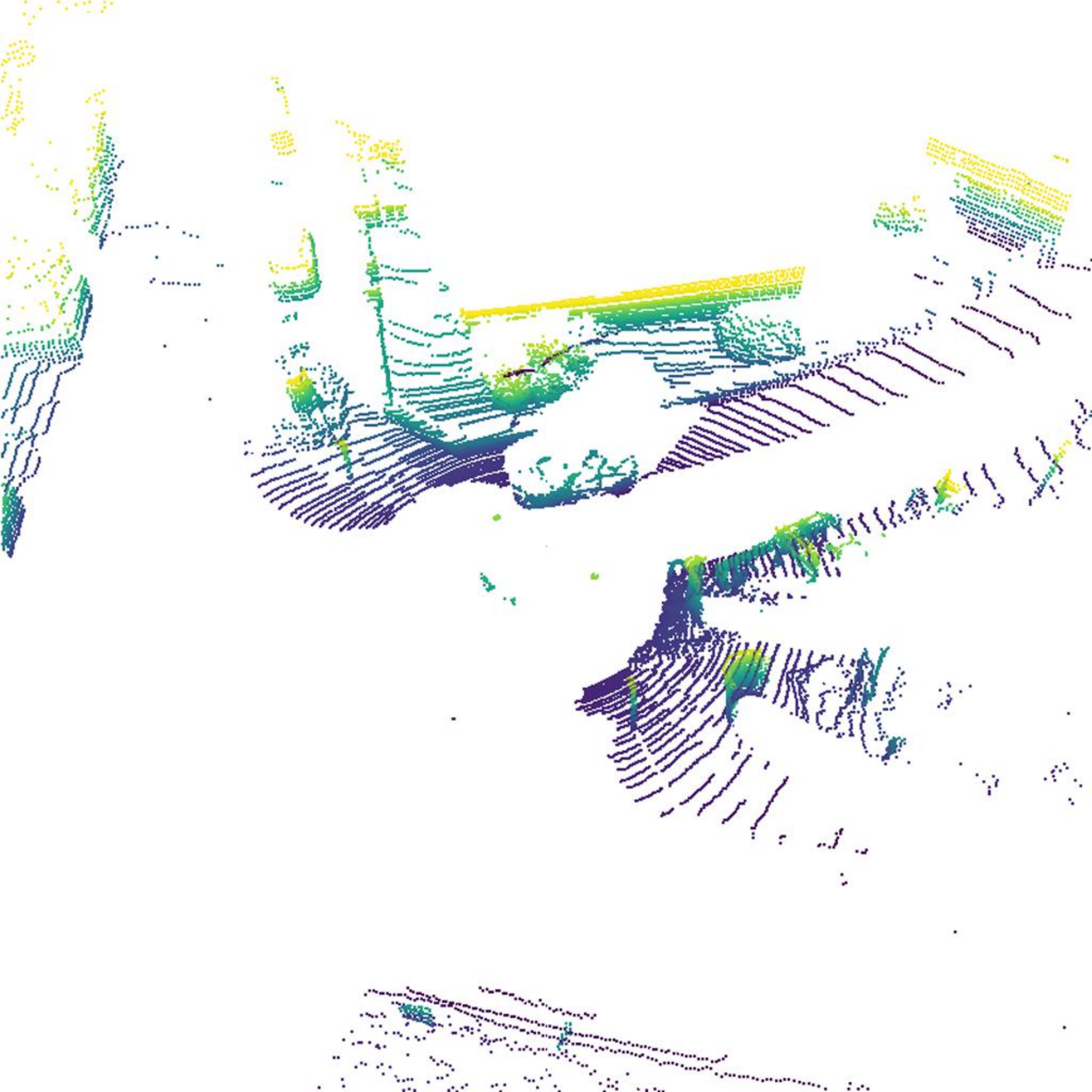}  \\
		Output & \includegraphics[align=c,width=\hsize]{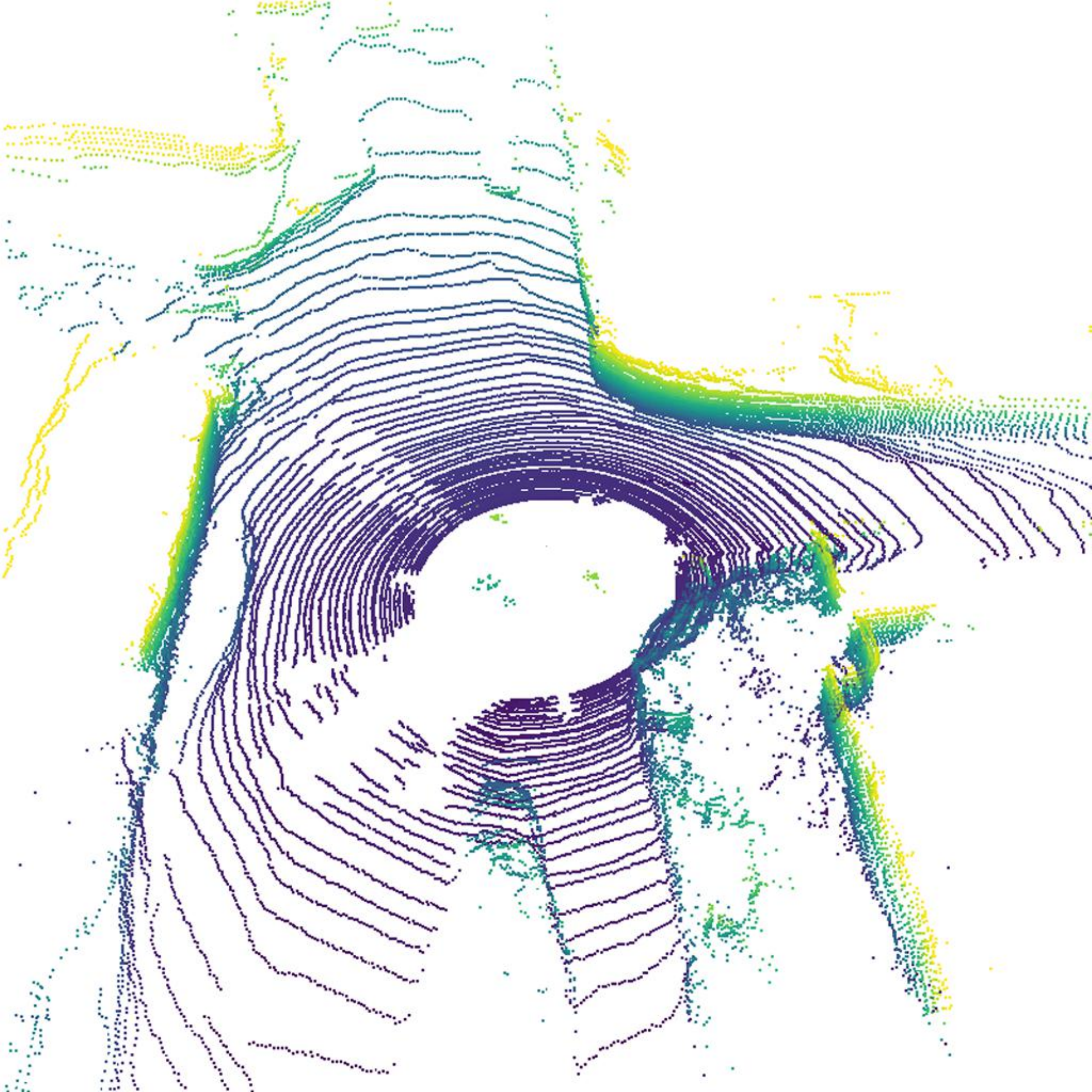} & \includegraphics[align=c,width=\hsize]{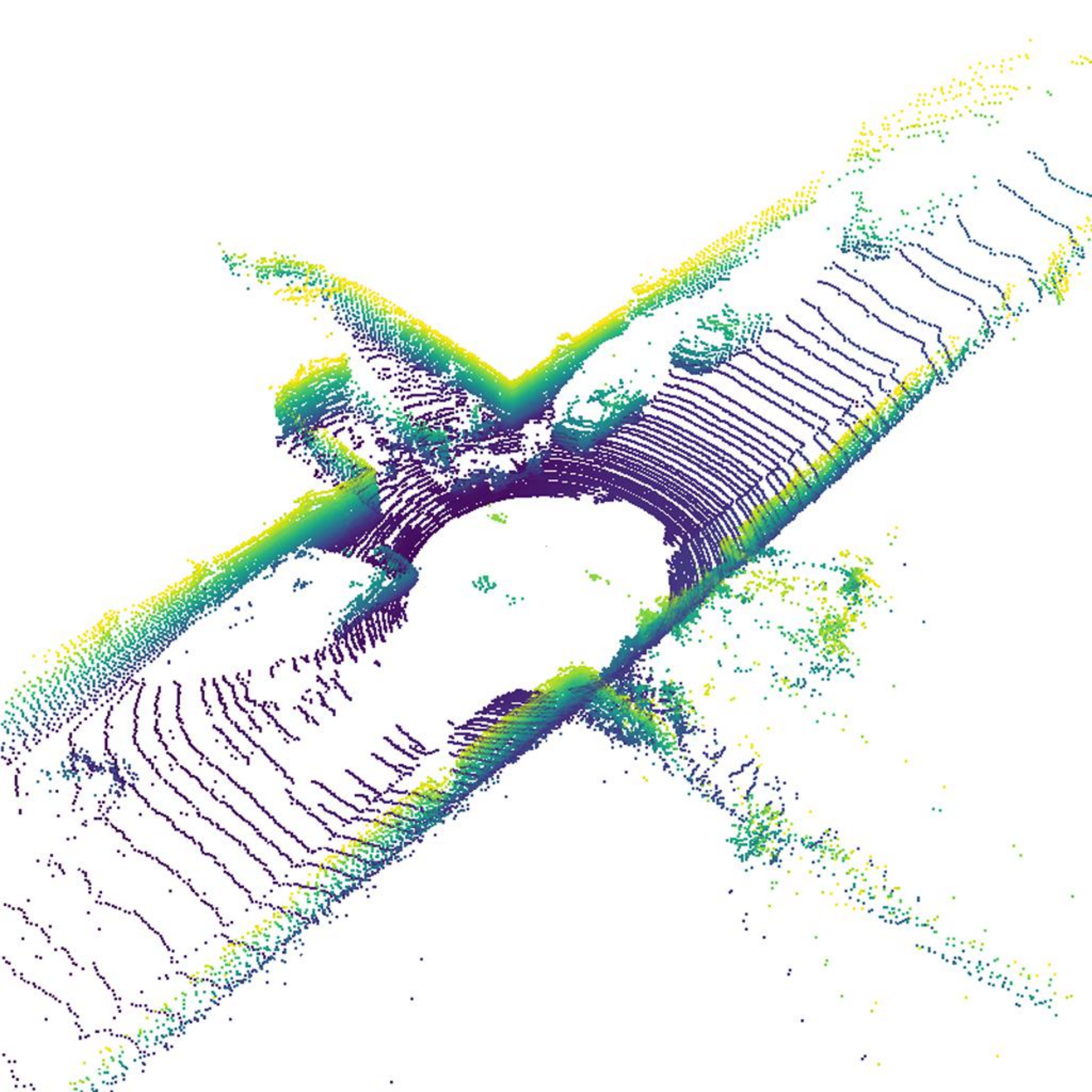} & \includegraphics[align=c,width=\hsize]{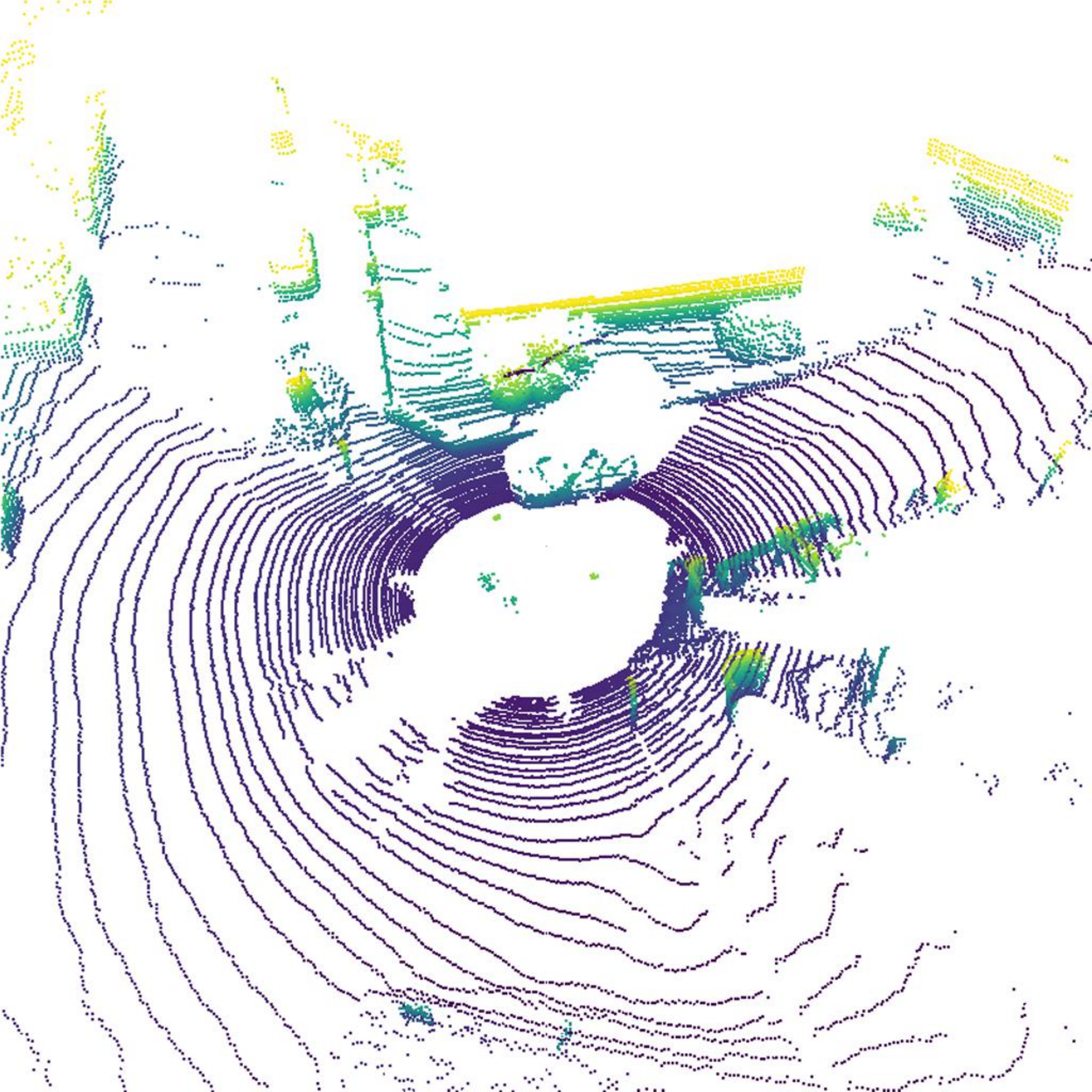} \\
		GT     & \includegraphics[align=c,width=\hsize]{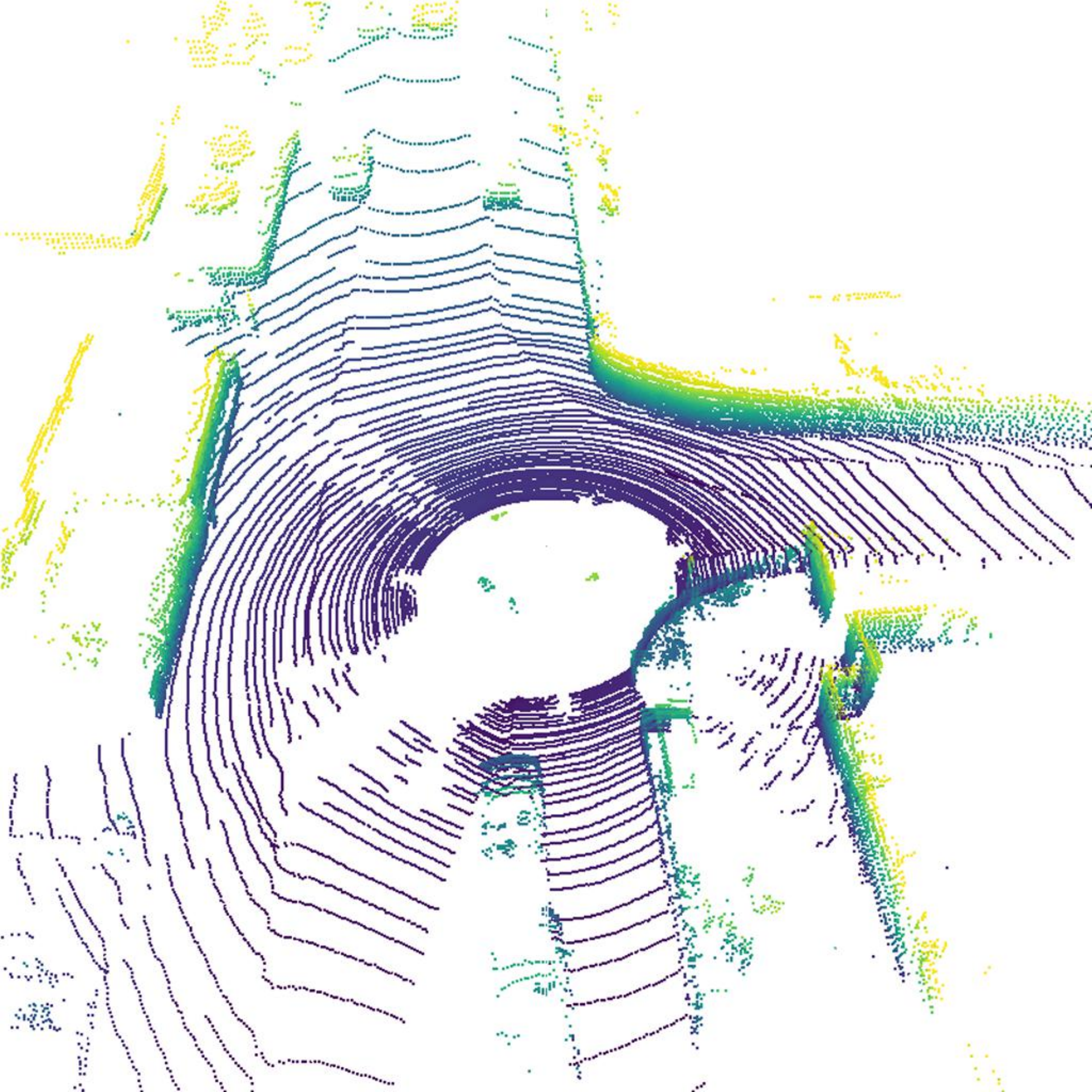}  & \includegraphics[align=c,width=\hsize]{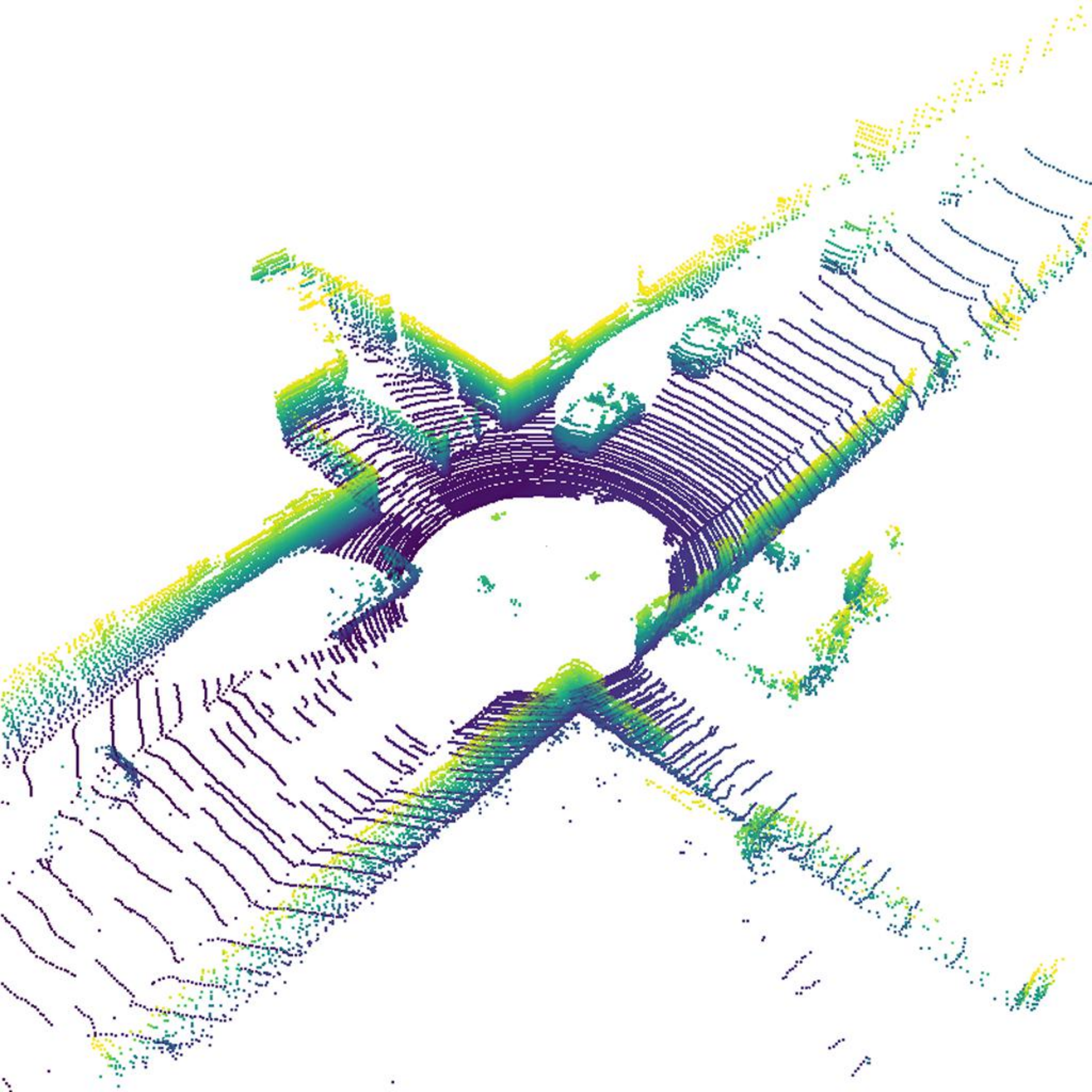}  & \includegraphics[align=c,width=\hsize]{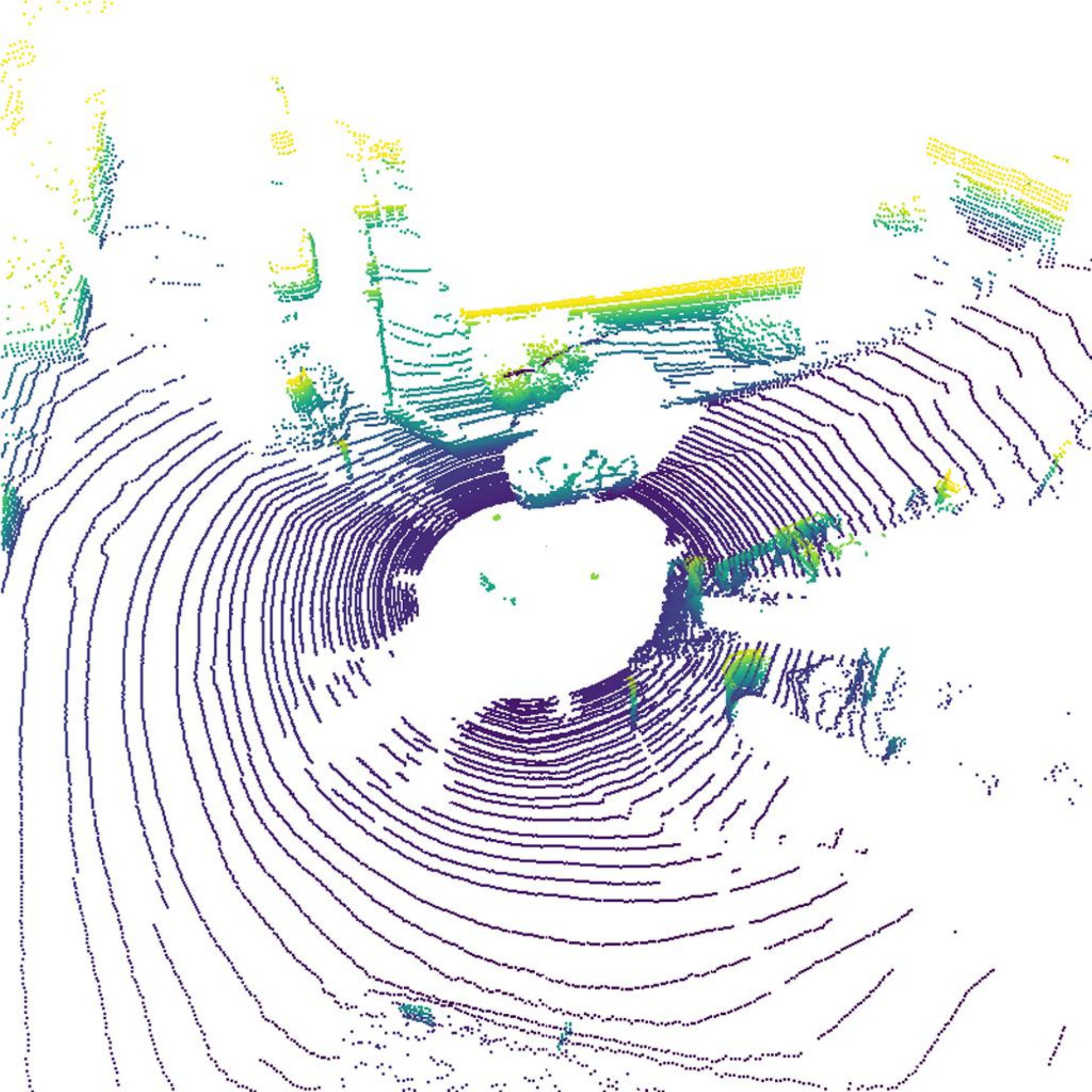}  
	\end{tabularx}
	\caption{\textbf{Conditional generation using R2DM}. We showcase the simulated corruptions (top) and our restored results (middle). Our method can handle the various levels of completion. The road removal (right) mimics the wet road situation.}
	\label{fig:conditional_generation_examples}
\end{figure}

Lugmayr~\etal~\cite{lugmayr2022repaint} proposed RePaint, an image inpainting method that exploits the pre-trained unconditional DDPMs.
In this work, we build a LiDAR completion pipeline by integrating our R2DM and RePaint.
In the sampling phase, $p(\bm{z}_s \mid \bm{z}_t)$ in Eq.~\ref{eq:p_zt_to_zs} follows by:
\begin{equation}
	\label{eq:repaint}
	\bm{z}_s = \bm{m} \odot \bm{z}_s^{\mathrm{known}} + (1-\bm{m}) \odot \bm{z}_s^{\mathrm{unknown}},
\end{equation}
where $\bm{m}\in\{0, 1\}^{{H}\times{W}}$ is a binary mask indicating known range/reflectance pixels, $\bm{z}_s^{\mathrm{known}}$ is \textit{noised} known pixels sampled by $q(\bm{z}_s \mid \bm{x})$ in Eq.~\ref{eq:q_x_to_zt}, and $\bm{z}_s^{\mathrm{unknown}}$ is \textit{denoised} unknown pixels sampled by $p(\bm{z}_s \mid \bm{z}_t)$ in  Eq.~\ref{eq:p_zt_to_zs}.
Moreover, to harmonize the gap between the known and unknown diffused extents, the sampling step above is repeatedly cycled with the forward diffusion step $q(\bm{z}_t \mid \bm{z}_s)$ in Eq.~\ref{eq:q_zs_to_zt}.
We set the number of sampling steps to 32 and the number of harmonization for each step to 10.
Fig.~\ref{fig:conditional_generation_examples} showcases our results on the beam-level, point-level, and object-level simulated corruptions.

\section{Experiments}
\label{sec:expeirments}

We evaluate our method on a generation task in Section~\ref{sec:generation_task} and a completion task in Section~\ref{sec:upsampling_task}.

\subsection{Generation task}
\label{sec:generation_task}

\subsubsection{Settings}

To evaluate performances on the \textit{unconditional} generation task, we use the KITTI-360 dataset~\cite{liao2022kitti}.
KITTI-360 consists of 81,106 point clouds measured by Velodyne HDL-64E (64-beam mechanical LiDAR sensor).
The data splits are from Zyrizanov~\etal~\cite{zyrianov2022learning}.
Each of the point clouds is projected to a $64\times1024$ image.
We compare different configurations of DDPM in terms of the loss functions, the range representations, and the positional encoding methods, as discussed in Section~\ref{sec:proposed_method}.
Table~\ref{tab:generation_kitti_360} lists a series of our training configurations (config A--H).
As the baseline, we compare with the state-of-the-art method, LiDARGen\footnote{We do not borrow the scores on the proceedings~\cite{zyrianov2022learning} but re-evaluate with the larger samples officially published at \url{https://github.com/vzyrianov/lidargen}.}~\cite{zyrianov2022learning}.
We also compare the GAN-based methods~\cite{caccia2019deep, nakashima2021learning, nakashima2023generative} using the KITTI-Raw dataset~\cite{geiger2013vision}.
Following the baselines, we employ the data splits from the Odometry benchmark~\cite{geiger2013vision} and the scan unfolding projection~\cite{triess2020scan}.
Note that, in this setting, the point cloud is first projected to a full resolution of $64\times2048$ and then subsampled to $64\times512$.

\begin{table*}[t]
	\centering
	\footnotesize
	\begin{threeparttable}
		\caption{Quantitative comparison of KITTI-360 generation.}
		\label{tab:generation_kitti_360}
		\begin{tabularx}{\hsize}{p{13mm}c ccCc CCCC}
			\toprule
			&              &       & \multicolumn{3}{c}{Configurations$^\ddag$} & Image               & Point cloud                     & \multicolumn{2}{c}{BEV}                                                                                          \\
			\cmidrule(lr){4-6} \cmidrule(lr){7-7} \cmidrule(lr){8-8} \cmidrule(lr){9-10}
			\multicolumn{2}{l}{Method (Framework)}                            & NFE          & Loss  & Range                                      & Positional encoding & FRD $\downarrow$                & FPD~$\downarrow$        & MMD{\tiny$\times10^4$~}$\downarrow$ & JSD{\tiny$\times10^2$~}$\downarrow$              \\
			\midrule
			\multicolumn{2}{l}{LiDARGen~(NCSNv2)~\cite{zyrianov2022learning}} & 1160$^\dag$  & $L_2$ & Log-scale                                  & Identity            & 579.39                          & \s90.29                 & \s7.39                              & \s7.38                                           \\
			\midrule
			\textbf{Ours}~(DDPM) & config A     & 256 & $L_2$             & Log-scale           & Identity                        & 202.40     & \s\s7.11     & \s1.67     & \s4.52     \\
			                     & config B     & 256 & \cc{gray!25}$L_1$ & Log-scale           & Identity                        & 382.35     & \s21.42      & \s7.70     & \s8.28     \\
			                     & config C     & 256 & \cc{gray!25}Huber & Log-scale           & Identity                        & 174.83     & \s11.20      & \s1.55     & \s4.71     \\
			                     & config D     & 256 & $L_2$             & \cc{gray!25}Metric  & Identity                        & 229.28     & \s12.03      & \s1.47     & \s4.01     \\
			                     & config E     & 256 & $L_2$             & \cc{gray!25}Inverse & Identity                        & 188.84     & \s19.66      & \s1.85     & \s3.12     \\
			                     & config F     & 256 & $L_2$             & Log-scale           & \cc{gray!25}w/o spatial bias    & 910.67     & 253.21       & 40.45      & 18.05      \\
			                     & config G     & 256 & $L_2$             & Log-scale           & \cc{gray!25}Spherical harmonics & 180.60     & \s\s4.90     & \s2.18     & \s4.12     \\
			                     & \bf config H & 256 & $L_2$             & Log-scale           & \cc{gray!25}Fourier features    & \bf 153.73 & \bf \s\s3.92 & \bf \s0.68 & \bf \s2.17 \\
			\bottomrule
		\end{tabularx}
		\begin{tablenotes}
			\item \dag~Five steps for each of the 232 noise levels. \ddag~The shaded cells indicate the differences from config A.
		\end{tablenotes}
	\end{threeparttable}
\end{table*}

\subsubsection{Implementation details}

Our models were trained in 300k steps while calculating an exponential moving average of model weights with a decay rate of 0.995 per every 10 steps.
We performed a distributed training with automatic mixed precision (AMP) on two NVIDIA A6000 GPUs.
Note that we did not use AMP for LiDARGen to avoid performance degradation.
Our model took approximately 20 GPU hours for training, and 30 GPU hours for generating 10k samples in 1024 denoising steps.
Our code and pre-trained weights are available at \url{https://github.com/kazuto1011/r2dm}.

\subsubsection{Evaluation metrics}

For the generation task, we compute three levels of distributional dissimilarity metrics employed in the LiDAR domain~\cite{zyrianov2022learning,nakashima2023generative}, between 10k generated samples and all available real samples.

\begin{itemize}
	\item \textit{Image-based}: Following the baseline~\cite{zyrianov2022learning}, we employ Fr\'echet range distance (FRD) for evaluating range and reflectance images.
	      The images are first fed into the off-the-shelf RangeNet-53~\cite{milioto2019rangenet} pre-trained a 19-class semantic segmentation task on SemanticKITTI~\cite{behley2019semantickitti}.
	      Then, FRD calculates the Fr\'echet distance~\cite{dowson1982frechet} between the generated and real sets on the feature space.
	\item \textit{Point cloud-based}: LiDAR range images can be transformed back to 3D point clouds.
	      Following the baseline~\cite{nakashima2023generative}, we also evaluate on the level of point clouds with the Fr\'echet point cloud distance (FPD)~\cite{shu20193d}.
	      FPD uses PointNet~\cite{qi2017pointnet1} pre-trained a 16-class classification task on ShapeNet~\cite{chang2015shapenet} and calculates the Fr\'echet distance on the feature space like FRD.
	\item \textit{BEV-based}: Following the baseline~\cite{zyrianov2022learning}, we also evaluate on the level of 2D bird's eye view (BEV)~\cite{zyrianov2022learning} projected from the point clouds.
	      We report the Jensen–Shannon divergence (JSD) and maximum-mean discrepancy (MMD), calculating the distance between the marginal point distributions on the 2D BEV histograms.
\end{itemize}

\subsubsection{Baseline results}

\begin{figure}[t]
	\centering
	\includegraphics[width=0.9\hsize]{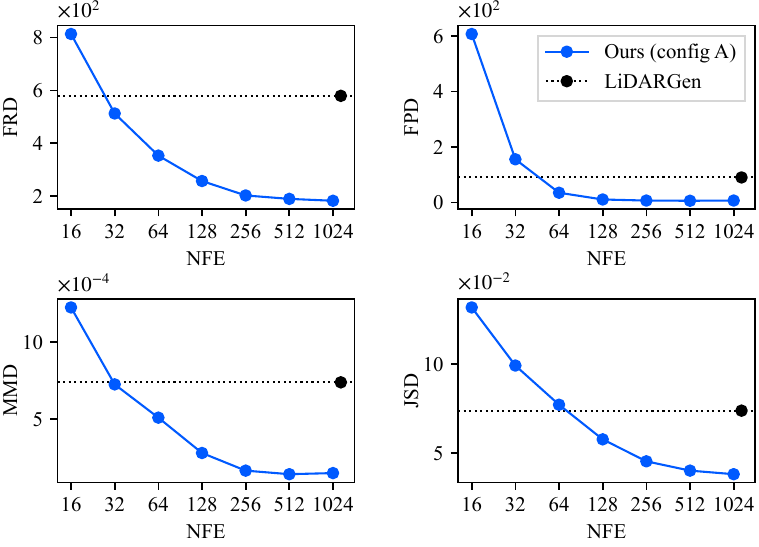}
	\caption{\textbf{Comparison of diffusion-based methods}. For overall metrics, \textcolor{blue}{our method} achieved better scores with the significantly lower number of function evaluations (NFE), against 1160 steps of LiDARGen~\cite{zyrianov2022learning}.}
	\label{fig:nfe_vs_scores}
\end{figure}

First, we compare LiDARGen~\cite{zyrianov2022learning} and our closest setup (DDPM config A).
Fig.~\ref{fig:nfe_vs_scores} shows the four evaluation scores as functions of the number of function evaluations (NFE)\footnote{NFE represents how many times the neural network is processed in sampling. For LiDARGen (NCSNv2), we count up the number of sampling steps for each noise level, which is 1160 steps in total.}.
For sampling of our DDPM, we swept $T$ in $\{16, 32, 64, 128, 256, 512, 1024\}$.
Intuitively, all scores of the DDPM improve as $T$ increases.
Moreover, the DDPM outperformed LiDARGen with significantly lower NFE.
Considering the tradeoff between computational time and quality, we will use $T=256$ for our DDPMs in the following experiments, unless otherwise specified.

\subsubsection{Ablations}

Comparing config A and B in Table~\ref{tab:generation_kitti_360}, we can see that $L_1$ loss is not effective for our settings.
We observed that samples of config B is overly smoothed due to the less sensitivity of $L_1$ loss.
Huber loss (config C), the combination of $L_1$ and $L_2$, showed the marginal improvements only on FRD and MMD.
Regarding the representation of the range modality, the metric depth (config D) and the inverse depth (config E) improved some metrics, while they harmed FPD.
Finally, our model with the positional encoding of Fourier features (config H) shows the best results with the large gaps from the other configurations.
Without the spatial bias (config F), DDPM degrades all the metrics, even worse than the baseline.
We hereinafter call config H as R2DM.

\subsubsection{Comparison to GANs}
\label{sec:comparison_to_gans}

Table~\ref{tab:generation_kitti_raw} compares our R2DM and GANs.
When increasing the number of timestep $T$ from the default setting ($T=256$), R2DM outperforms the baseline in FPD.
We believe that the performance gap with the KITTI-360 experiment lies in the setup of range images.
In KITTI-360 experiments, the range images were downscaled to alleviate missing points called ray-drop noises.
In contrast, the range images of KITTI-Raw were also downscaled but the ray-drop noises were retained to be closer to raw scan data.
It is considered that there is room for further ingenuity to handle noisy settings, such as full resolution.

\begin{table}[t]
	\centering
	\scriptsize
	\begin{threeparttable}
		\caption{Quantitative comparison on KITTI-Raw generation.}
		\label{tab:generation_kitti_raw}
		\begin{tabularx}{\hsize}{l CcCC}
			\toprule
			& Image             & Point cloud      & \multicolumn{2}{c}{BEV}                                                     \\
			\cmidrule(lr){2-2} \cmidrule(lr){3-3} \cmidrule(lr){4-5}
			Method                                                  & FRD~$\downarrow$  & FPD~$\downarrow$ & MMD{\tiny$\times10^4$~}$\downarrow$ & JSD{\tiny$\times10^2$~}$\downarrow$ \\
			\midrule
			Vanilla GAN~\cite{caccia2019deep,nakashima2021learning} & \tc{gray!75}{N/A} & 3657.60          & 1.02                                & 5.03                                \\
			DUSty v1~\cite{nakashima2021learning}                   & \tc{gray!75}{N/A} & \s223.63         & 0.80                                & 2.87                                \\
			DUSty v2~\cite{nakashima2023generative}                 & \tc{gray!75}{N/A} & \s\s98.02        & \bf 0.22                            & \bf 2.86                            \\
			\midrule
			\textbf{R2DM ($T=256$)}                                 & 215.27            & \s128.74         & 0.72                                & 3.79                                \\
			\textbf{R2DM ($T=512$)}                                 & 209.24            & \s\s89.62        & 0.65                                & 3.76                                \\
			\textbf{R2DM ($T=1024$)}                                & \bf 207.31        & \bf \s\s70.34    & 0.44                                & 3.56                                \\
			\bottomrule
		\end{tabularx}
		\begin{tablenotes}
			\item FRD is not available for the baselines~\cite{nakashima2021learning,nakashima2023generative} which do not support the reflectance.
		\end{tablenotes}
	\end{threeparttable}
\end{table}

\begin{figure*}[t]
	\centering
	\scriptsize
	\begin{tabularx}{\hsize}{cCCC}
		\rotatebox[origin=c]{90}{KITTI-360~\cite{liao2022kitti}} &
		\includegraphics[align=c,width=\hsize]{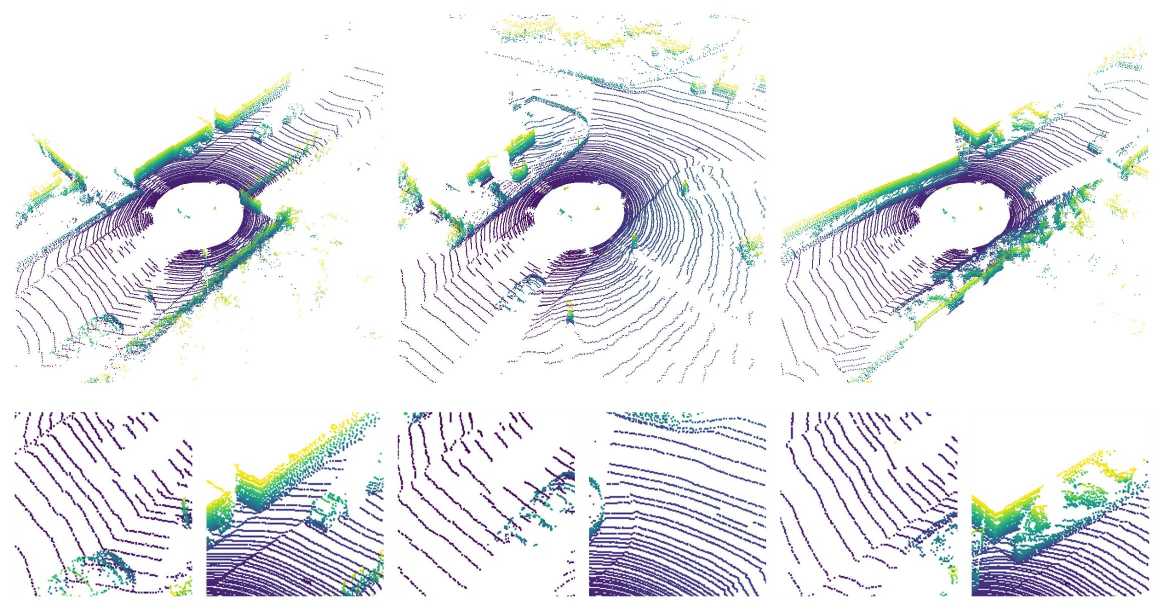}    &
		\includegraphics[align=c,width=\hsize]{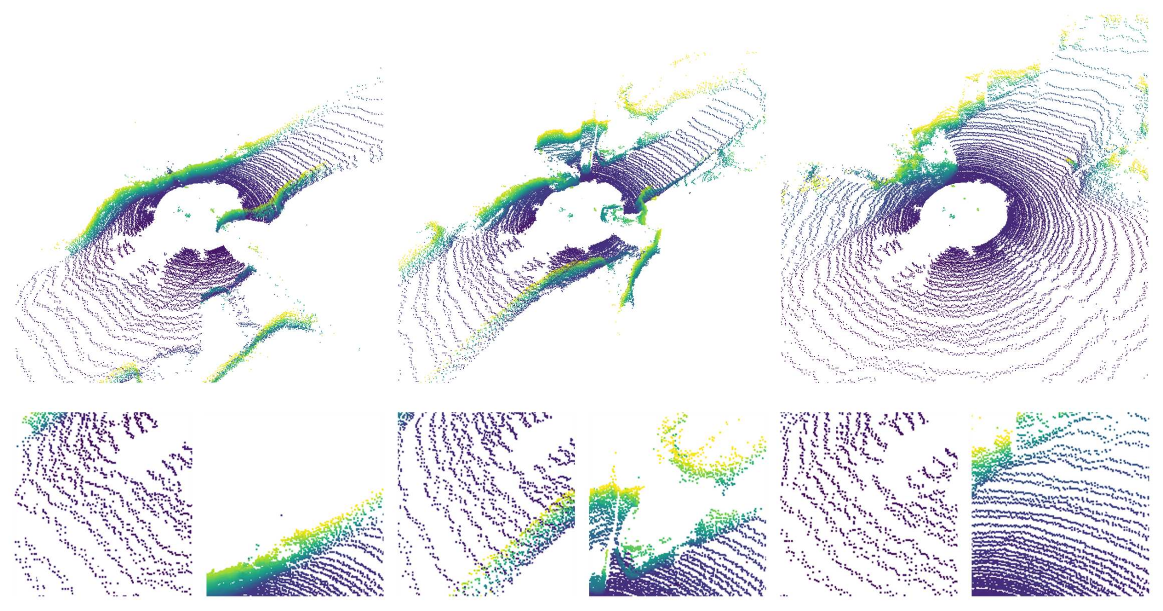} &
		\includegraphics[align=c,width=\hsize]{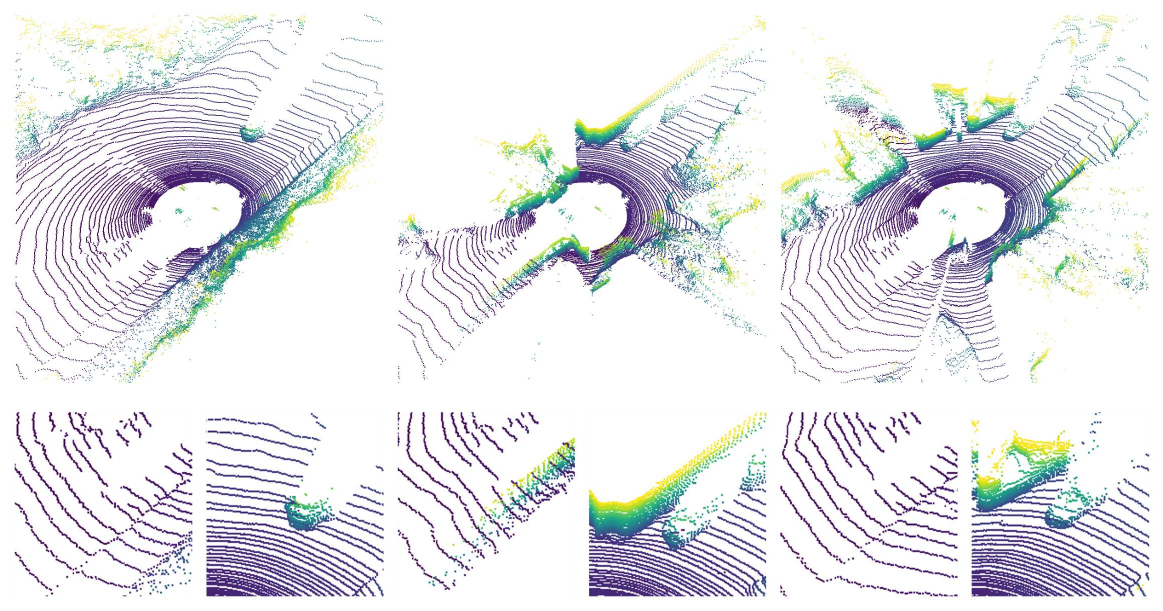}                                                                          \\
		\cmidrule(lr){2-2} \cmidrule(lr){3-3} \cmidrule(lr){4-4}
		  & Training data & Baseline (LiDARGen~\cite{zyrianov2022learning}) & \textbf{R2DM (ours)} 
	\end{tabularx}
	\\ \vspace{3mm}
	\begin{tabularx}{\hsize}{cCCC}
		\rotatebox[origin=c]{90}{KITTI-Raw~\cite{geiger2013vision}} &
		\includegraphics[align=c,width=\hsize]{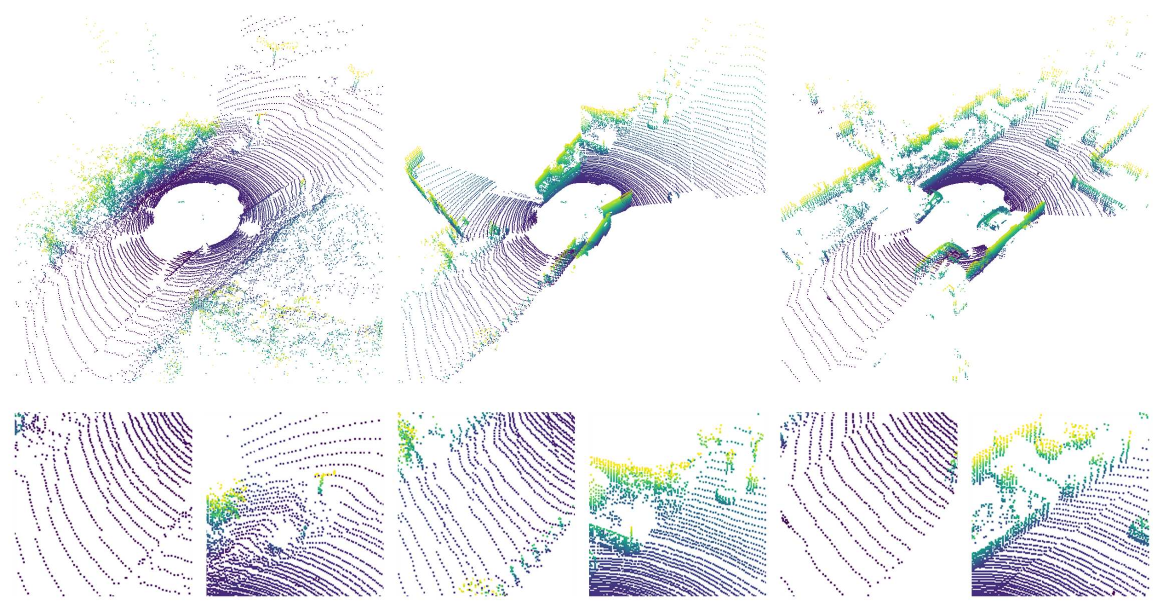}   &
		\includegraphics[align=c,width=\hsize]{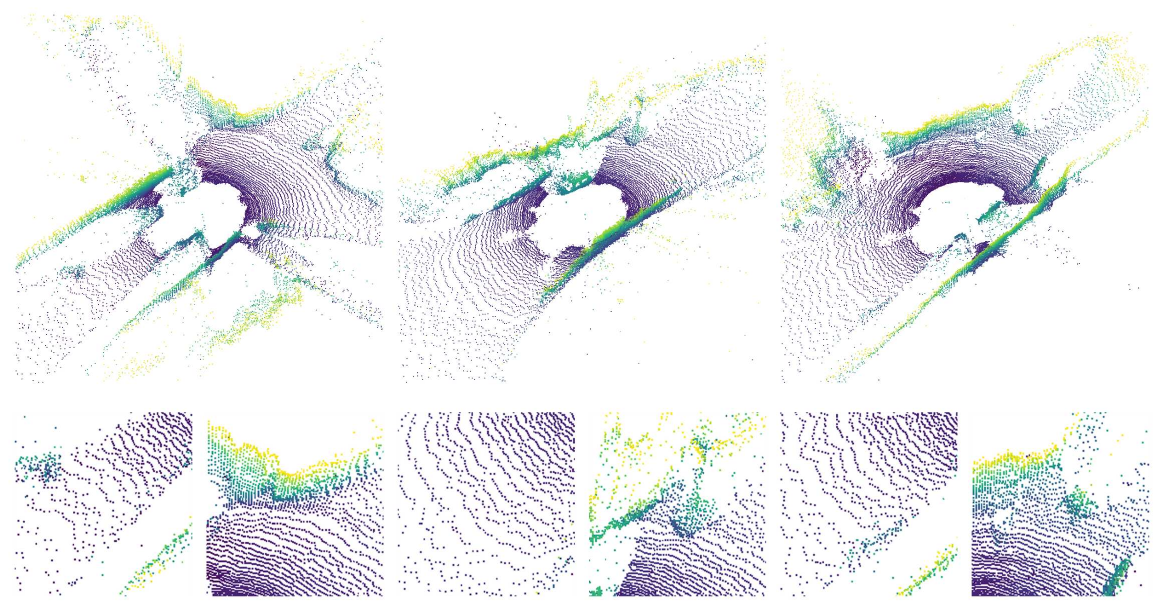} &
		\includegraphics[align=c,width=\hsize]{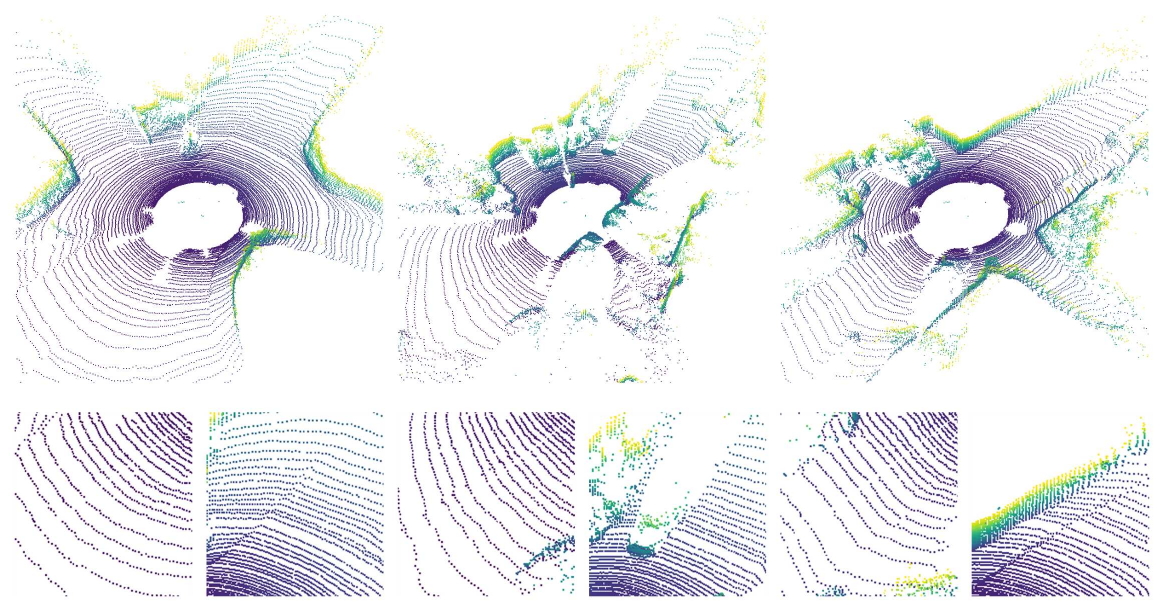}                                                                            \\
		\cmidrule(lr){2-2} \cmidrule(lr){3-3} \cmidrule(lr){4-4}
		  & Training data & Baseline (DUSty v2~\cite{nakashima2023generative}) & \textbf{R2DM (ours)} 
	\end{tabularx}
	\\ \vspace{3mm}
	\caption{\textbf{Unconditional generation results}. We show the results of the baselines and our method on KITTI-360 (top) and KITTI-Raw (bottom). The LiDARGen results are from officially released samples~\cite{zyrianov2022learning}. The DUSty v2 results are generated using the official pre-trained models~\cite{nakashima2023generative}.}
	\label{fig:unconditional_generation}
\end{figure*}

\subsubsection{Qualitative results}

Fig.~\ref{fig:unconditional_generation} shows the generated samples of KITTI-360 and KITTI-Raw.
We compare the real data, the baseline, and our method.
We can see that our method realizes high-fidelity structures of LiDAR point clouds both locally and globally.
For instance, regarding the scan lines (zoomed regions in Fig.~\ref{fig:unconditional_generation}), our results are less noisy and close to the real data compared to the baselines.

\subsection{Completion task}
\label{sec:upsampling_task}

\subsubsection{Settings}

Following LiDARGen~\cite{zyrianov2022learning}, we conduct a beam-level $4\times$ upsampling experiment on KITTI-360 to assess the completion method described in Section~\ref{sec:lidar_completion}.
We simulate the sparse input by subsampling every four horizontal scan lines.
We compare our method with simple interpolation methods (nearest-neighbor, bilinear, bicubic), simulation-based supervised methods (LiDAR-SR~\cite{shan2020simulation}, ILN~\cite{kwon2022implicit}) trained on the CARLA simulator~\cite{dosovitskiy2017carla}, and the diffusion model-based approach LiDARGen~\cite{zyrianov2022learning}.
LiDARGen leverages posterior sampling of score-based generative models, that is, guiding the scores in Langevin dynamics based on the errors of known pixels.

\subsubsection{Quantitative results}

To evaluate point-wise reconstruction performance, we compute mean absolute error (MAE) for both range and reflectance modalities.
Furthermore, we evaluate semantic consistency by calculating the intersection-over-union (IoU, \%) between the segmentation labels of upsampled and ground truth data. These labels are predicted using the pre-trained RangeNet-53 model~\cite{milioto2019rangenet}.
We use ground-truth reflectance as an upper bound for the simulation-based methods, which do not support reflectance modality.
The evaluation results are shown in Table~\ref{tab:completion_kitti_360}.
Our R2DM outperforms the baselines for all evaluation metrics with large margins.

\begin{table}[t]
	\centering
	\scriptsize
	\begin{threeparttable}
		\caption{Beam-level $4\times$ upsampling on KITTI-360 test set.}
		\label{tab:completion_kitti_360}
		\begin{tabularx}{\hsize}{l l C c C}
			\toprule
			                               &                                                   & Range            & Reflectance       & Semantics                \\
			\cmidrule(lr){3-3} \cmidrule(lr){4-4} \cmidrule(lr){5-5}
			Approach                       & Method                                            & MAE~$\downarrow$ & MAE~$\downarrow$  & IoU{\tiny~\%}~$\uparrow$ \\
			\midrule
			\multirow{3}{*}{Interpolation} & Nearest-neighbor                                  & 2.083            & 0.106             & 18.78                    \\
			                               & Bilinear                                          & 2.110            & 0.101             & 18.17                    \\
			                               & Bicubic                                           & 2.297            & 0.108             & 18.54                    \\
			\midrule
			\multirow{2}{*}{Supervised}    & LiDAR-SR~\cite{shan2020simulation}                & 2.085            & \tc{gray!75}{N/A} & 21.61                    \\
			                               & ILN~\cite{kwon2022implicit}                       & 2.237            & \tc{gray!75}{N/A} & 21.80                    \\
			\midrule
			\multirow{2}{*}{Diffusion}     & LiDARGen~\cite{zyrianov2022learning}              & 1.551            & 0.080             & 22.46                    \\
			                               & \textbf{R2DM} + RePaint~\cite{lugmayr2022repaint} & \bf 0.923        & \bf 0.050         & \bf 34.44                \\
			\bottomrule
		\end{tabularx}
	\end{threeparttable}
\end{table}

\subsubsection{Qualitative results}

\begin{figure}[t]
	\centering
	\scriptsize
	\begin{tabularx}{\hsize}{CCCC}
		Sparse input                                                                  & LiDARGen~\cite{zyrianov2022learning}                                               & \textbf{R2DM (ours)}                                                           & GT                                                                           \\
		\cmidrule(lr){1-1} \cmidrule(lr){2-2} \cmidrule(lr){3-3} \cmidrule(lr){4-4}
		\includegraphics[align=c,width=\hsize]{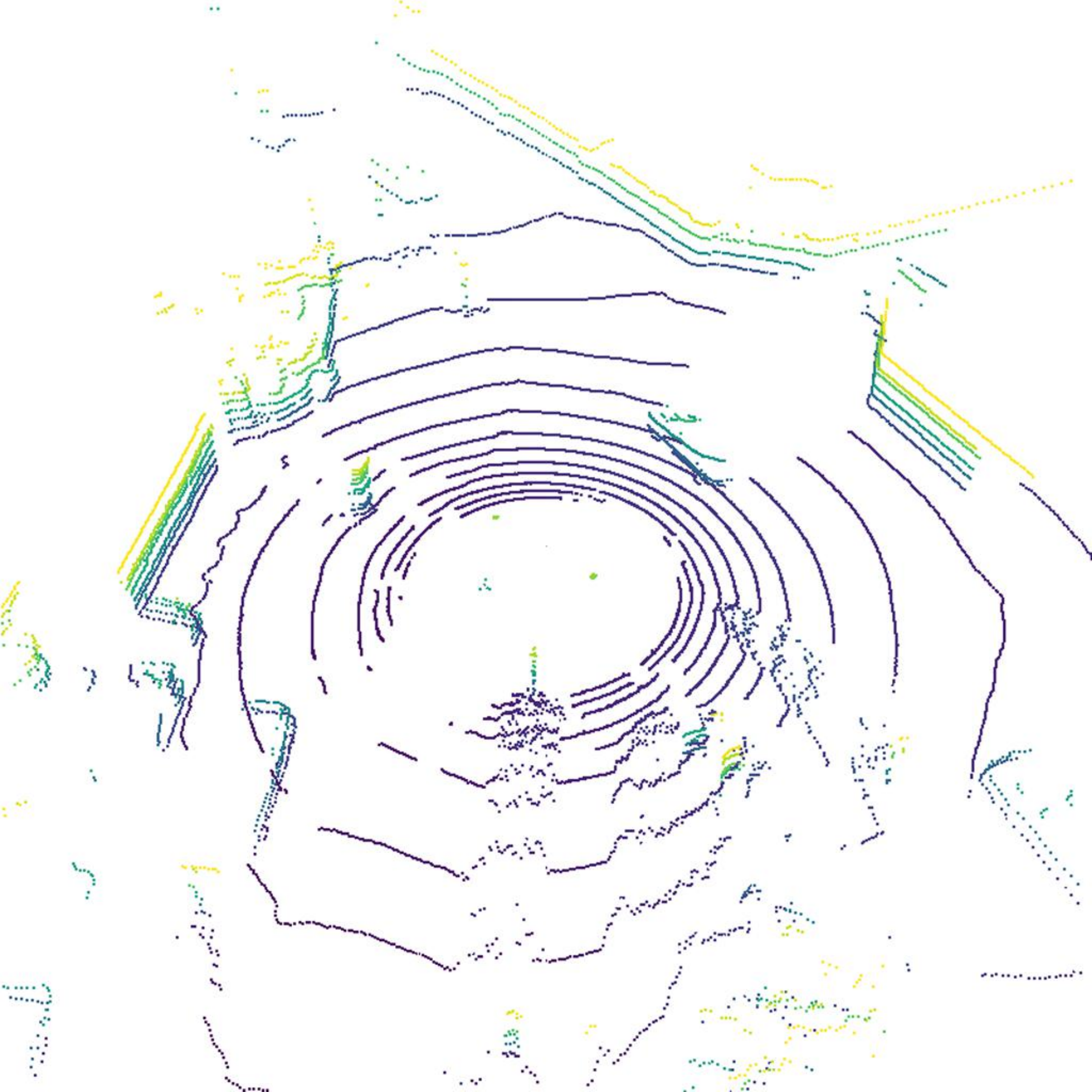} & \includegraphics[align=c,width=\hsize]{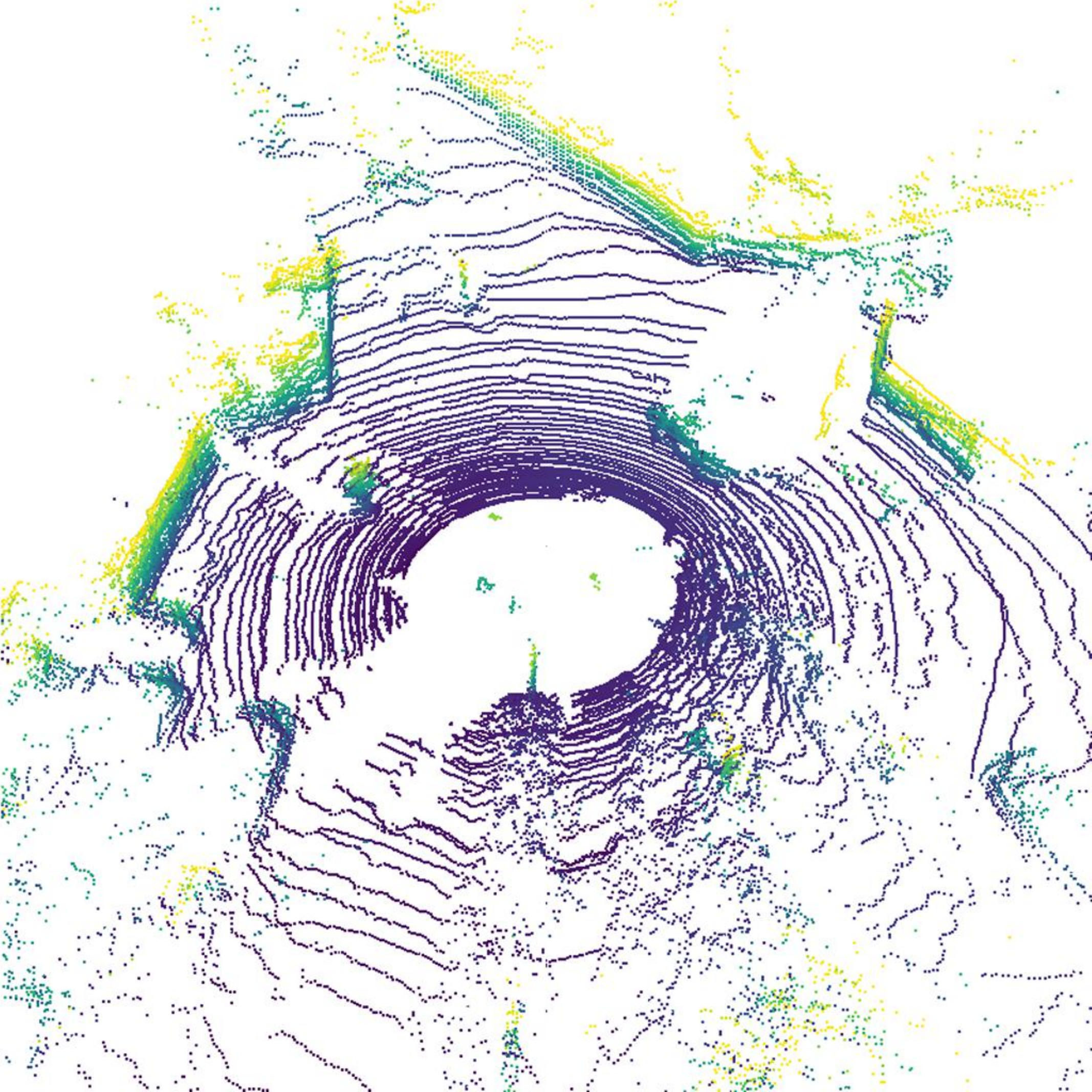}    & \includegraphics[align=c,width=\hsize]{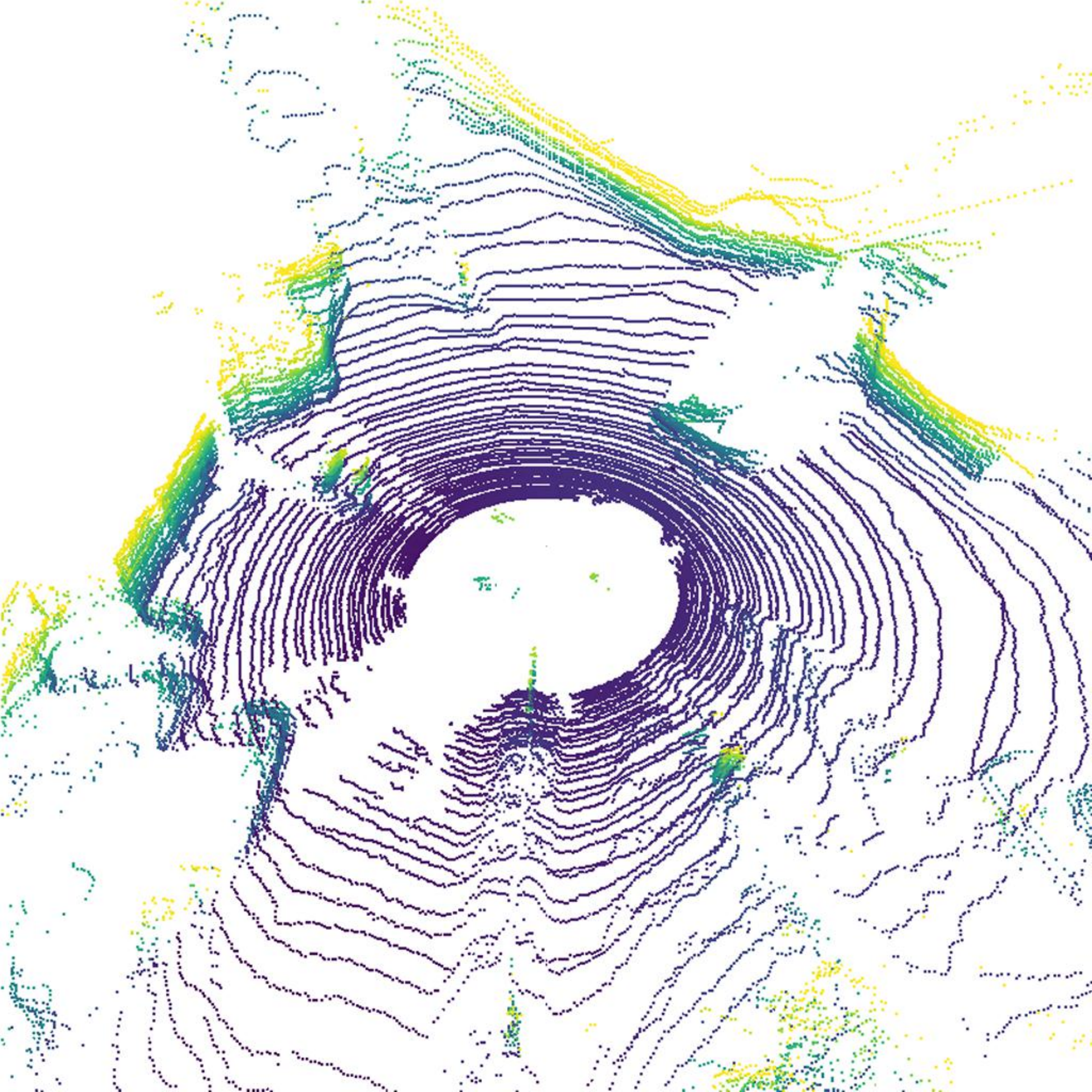}    & \includegraphics[align=c,width=\hsize]{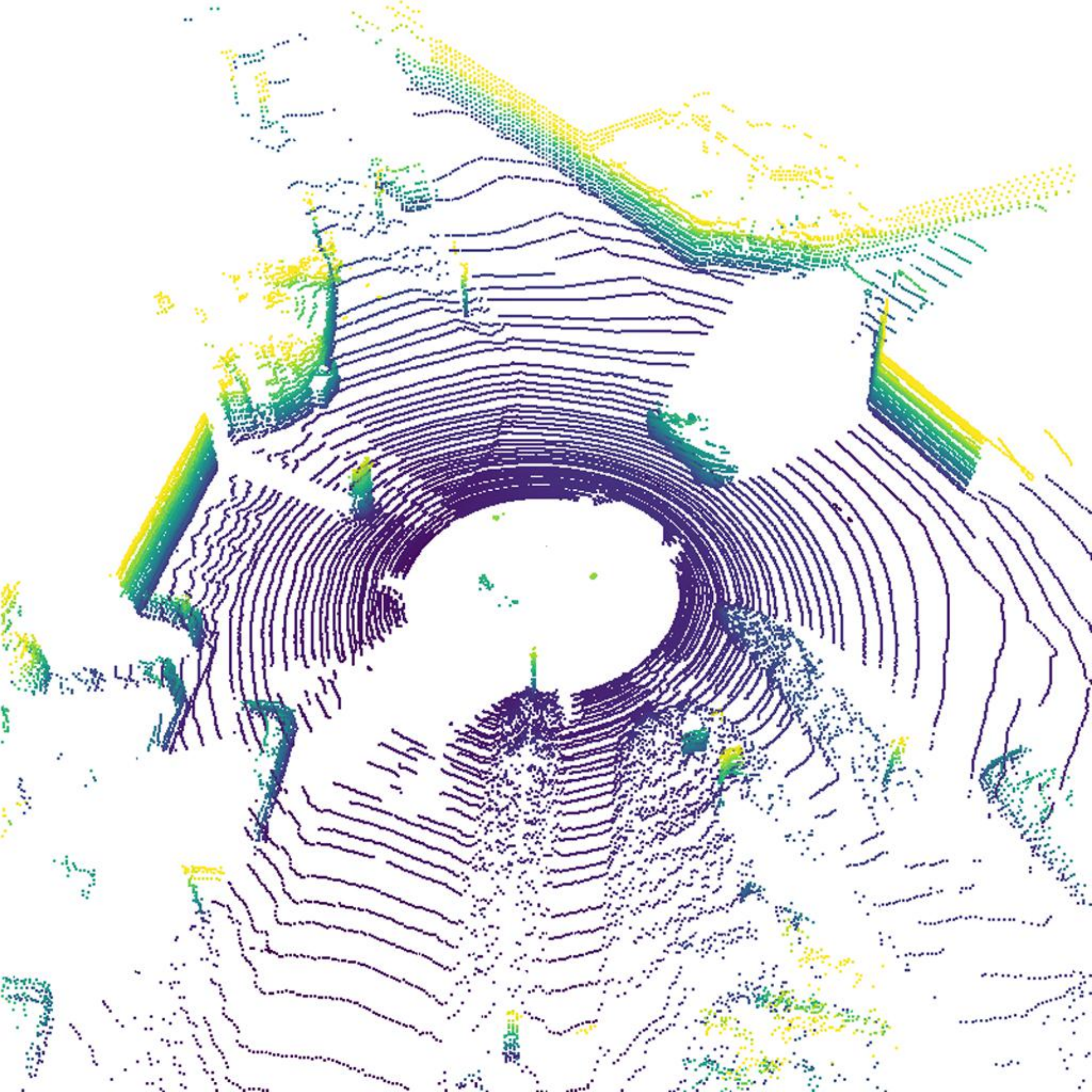}    \\
		                                                                              & \includegraphics[align=c,width=\hsize]{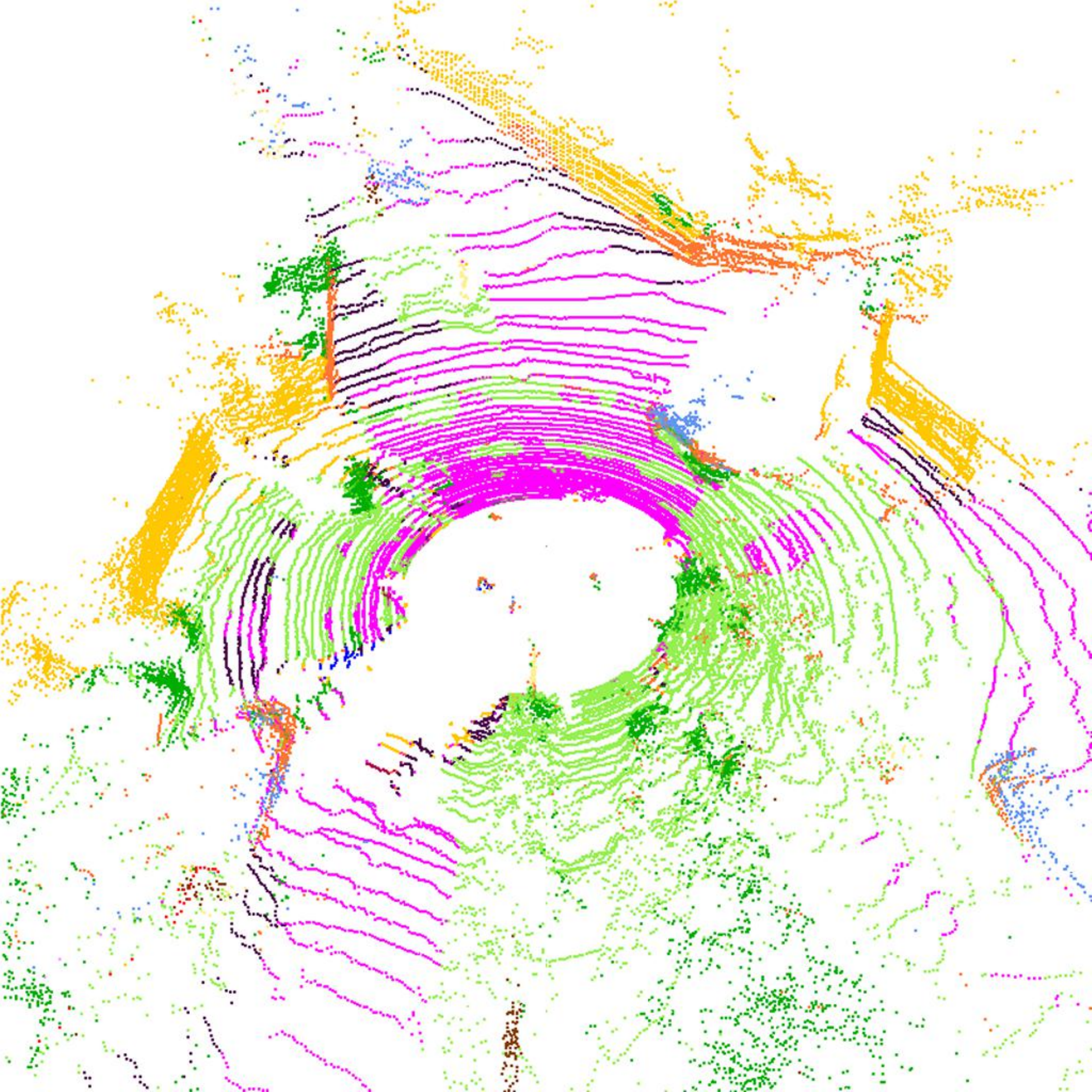} & \includegraphics[align=c,width=\hsize]{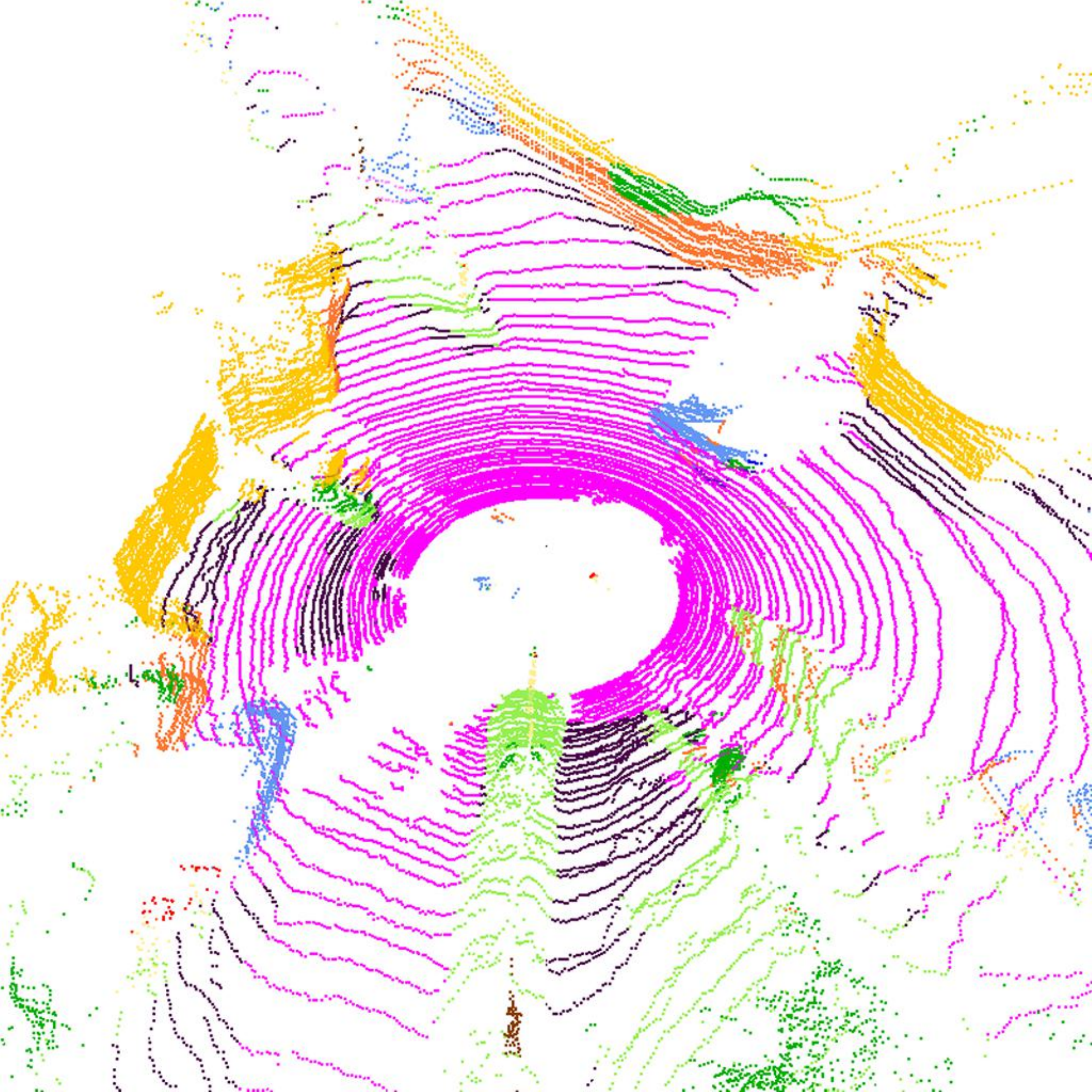} & \includegraphics[align=c,width=\hsize]{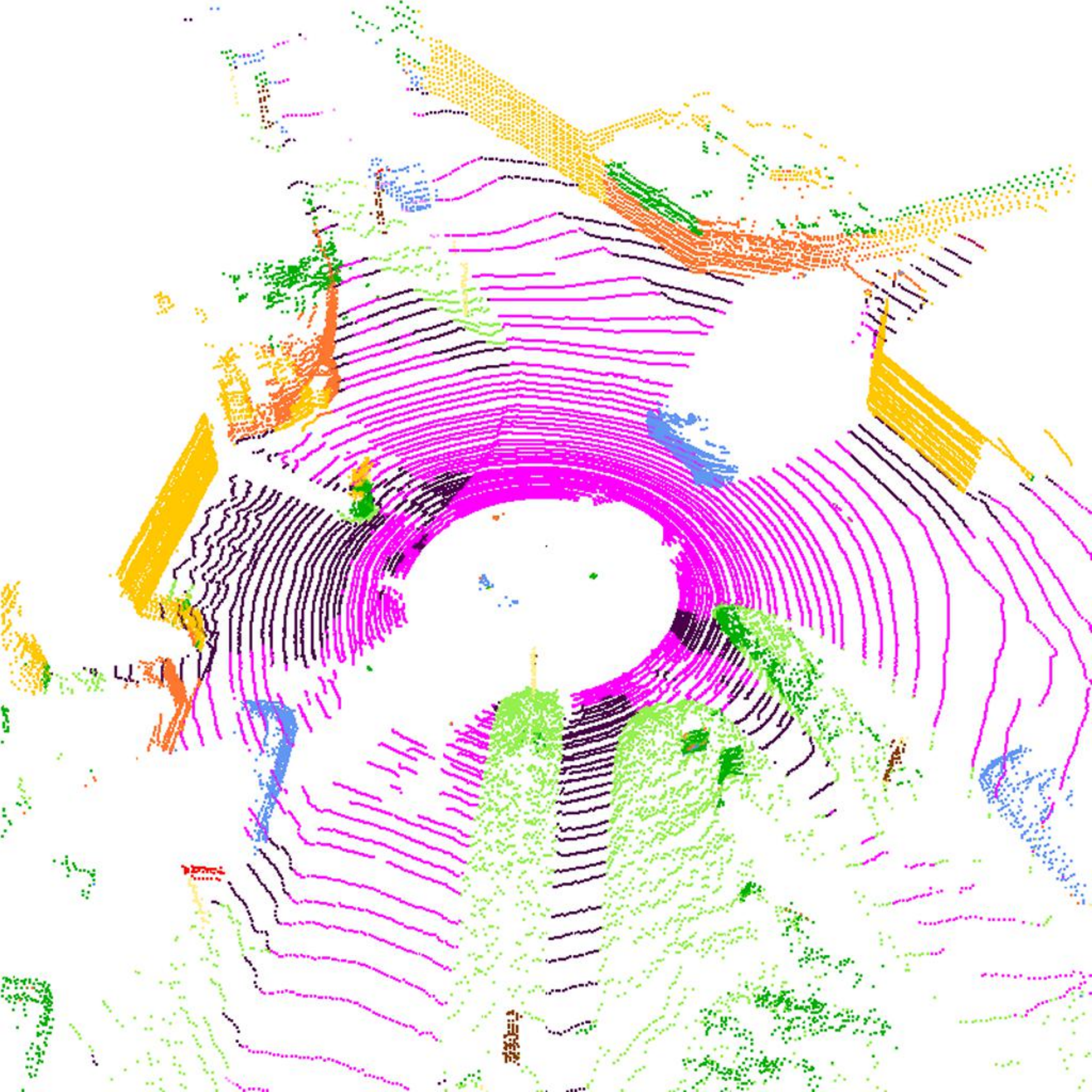} \\
	\end{tabularx}
	\caption{\textbf{Beam-level upsampling results on KITTI-360}. We compare the upsampled point clouds (top) and the semantic segmentation results (bottom) between the diffusion-based methods. The semantic segmentation results are color-coded according to the predicted classes.}
	\label{fig:completion_and_segmentation_kitti_360}
\end{figure}

We provide the results of upsampling and semantic segmentation in Fig.~\ref{fig:completion_and_segmentation_kitti_360}.
Compared to the baseline, our method shows better fidelity and consistency on the point clouds and the predicted classes.

\section{Conclusions}

We proposed R2DM, a denoising diffusion probabilistic model for realistic LiDAR range/reflectance generation based on the image representation.
R2DM achieved the state-of-the-art performance with the lower computational cost against the baseline.
Our exploration of the effective model revealed that the fidelity in point clouds can be significantly improved by introducing the explicit spatial bias with Fourier features.
Furthermore, we proposed a completion pipeline that leveraged the pre-trained R2DM and demonstrated the beam-level upsampling task.
Future work involves investigating model scalability, noise-robust training, and further applications to perception tasks.

\bibliographystyle{ieeetr}
\bibliography{main}

\end{document}

%% file: preamble.tex
\usepackage{graphicx}
\usepackage{amsmath}
\usepackage{amssymb}
\usepackage[space]{cite}
\usepackage{bm}
\usepackage{booktabs}
\usepackage{tabularx}
\usepackage{color}
\usepackage{url}
\usepackage{multirow}
\usepackage{gensymb}
\usepackage{dsfont}
\usepackage{xcolor}
\usepackage{colortbl}
\usepackage{url}
\usepackage{siunitx}
\usepackage{threeparttable}
\usepackage{graphbox}

\makeatletter
\let\NAT@parse\undefined
\makeatother
\usepackage{hyperref}
\hypersetup{
	bookmarks         = true,
	bookmarkstype     = toc,
	bookmarksnumbered = true,
	bookmarksopen     = true,
	colorlinks        = false,
	hidelinks,
	breaklinks        = false,
}

\newcolumntype{C}{>{\centering\arraybackslash}X}
\newcolumntype{L}{>{\raggedright\arraybackslash}X}
\newcolumntype{R}{>{\raggedleft\arraybackslash}X}
\newcolumntype{P}[1]{>{\centering\arraybackslash}p{#1}}

\newcommand{\etal}{\textit{et al.}}
\newcommand{\s}{\hphantom{0}}
\DeclareRobustCommand{\cc}[1]{{\cellcolor{#1}}}
\DeclareRobustCommand{\tc}[2]{{\textcolor{#1}{#2}}}

%% file: main.bbl
\begin{thebibliography}{10}

\bibitem{li2020deep}
Y.~Li, L.~Ma, Z.~Zhong, F.~Liu, M.~A. Chapman, D.~Cao, and J.~Li, ``Deep learning for {LiDAR} point clouds in autonomous driving: A review,'' {\em IEEE Transactions on Neural Networks and Learning Systems}, vol.~32, no.~8, pp.~3412--3432, 2020.

\bibitem{bond2022deep}
S.~Bond-Taylor, A.~Leach, Y.~Long, and C.~G. Willcocks, ``Deep generative modelling: A comparative review of {VAE}s, {GAN}s, normalizing flows, energy-based and autoregressive models,'' {\em IEEE Transactions on Pattern Analysis and Machine Intelligence (TPAMI)}, vol.~44, no.~11, pp.~7327--7347, 2022.

\bibitem{caccia2019deep}
L.~Caccia, H.~van Hoof, A.~Courville, and J.~Pineau, ``Deep generative modeling of {LiDAR} data,'' in {\em Proceedings of the IEEE/RSJ International Conference on Intelligent Robots and Systems (IROS)}, pp.~5034--5040, 2019.

\bibitem{nakashima2021learning}
K.~Nakashima and R.~Kurazume, ``Learning to drop points for {LiDAR} scan synthesis,'' in {\em Proceedings of the IEEE/RSJ International Conference on Intelligent Robots and Systems (IROS)}, pp.~222--229, 2021.

\bibitem{nakashima2023generative}
K.~Nakashima, Y.~Iwashita, and R.~Kurazume, ``Generative range imaging for learning scene priors of {3D} {LiDAR} data,'' in {\em Proceedings of the IEEE/CVF Winter Conference on Applications of Computer Vision (WACV)}, pp.~1256--1266, 2023.

\bibitem{zyrianov2022learning}
V.~Zyrianov, X.~Zhu, and S.~Wang, ``Learning to generate realistic {LiDAR} point clouds,'' in {\em Proceedings of the European Conference on Computer Vision (ECCV)}, pp.~17--35, 2022.

\bibitem{kingma2014auto}
D.~P. Kingma and M.~Welling, ``Auto-encoding variational bayes,'' in {\em Proceedings of the International Conference on Learning Representations (ICLR)}, 2014.

\bibitem{goodfellow2014generative}
I.~Goodfellow, J.~Pouget-Abadie, M.~Mirza, B.~Xu, D.~Warde-Farley, S.~Ozair, A.~Courville, and Y.~Bengio, ``Generative adversarial nets,'' in {\em Proceedings of the Advances in Neural Information Processing Systems (NeurIPS)}, pp.~2672--2680, 2014.

\bibitem{song2019generative}
Y.~Song and S.~Ermon, ``Generative modeling by estimating gradients of the data distribution,'' in {\em Proceedings of the Advances in neural information processing systems (NeurIPS)}, pp.~11895--11907, 2019.

\bibitem{song2020improved}
Y.~Song and S.~Ermon, ``Improved techniques for training score-based generative models,'' in {\em Proceedings of the Advances in neural information processing systems (NeurIPS)}, vol.~33, pp.~12438--12448, 2020.

\bibitem{song2021scorebased}
Y.~Song, J.~Sohl-Dickstein, D.~P. Kingma, A.~Kumar, S.~Ermon, and B.~Poole, ``Score-based generative modeling through stochastic differential equations,'' in {\em Proceedings of the International Conference on Learning Representations (ICLR)}, 2021.

\bibitem{ho2020denoising}
J.~Ho, A.~Jain, and P.~Abbeel, ``Denoising diffusion probabilistic models,'' in {\em Proceedings of the Advances in Neural Information Processing Systems (NeurIPS)}, vol.~33, pp.~6840--6851, 2020.

\bibitem{nichol2021improved}
A.~Q. Nichol and P.~Dhariwal, ``Improved denoising diffusion probabilistic models,'' in {\em Proceedings of the International Conference on Machine Learning (ICML)}, pp.~8162--8171, 2021.

\bibitem{kingma2021variational}
D.~Kingma, T.~Salimans, B.~Poole, and J.~Ho, ``Variational diffusion models,'' in {\em Proceedings of the Advances in neural information processing systems (NeurIPS)}, vol.~34, pp.~21696--21707, 2021.

\bibitem{saharia2022photorealistic}
C.~Saharia, W.~Chan, S.~Saxena, L.~Li, J.~Whang, E.~L. Denton, K.~Ghasemipour, R.~Gontijo~Lopes, B.~Karagol~Ayan, T.~Salimans, {\em et~al.}, ``Photorealistic text-to-image diffusion models with deep language understanding,'' in {\em Proceedings of the Advances in Neural Information Processing Systems (NeurIPS)}, vol.~35, pp.~36479--36494, 2022.

\bibitem{liao2022kitti}
Y.~Liao, J.~Xie, and A.~Geiger, ``{KITTI-360}: A novel dataset and benchmarks for urban scene understanding in {2D} and {3D},'' {\em IEEE Transactions on Pattern Analysis and Machine Intelligence (TPAMI)}, vol.~45, no.~3, pp.~3292--3310, 2022.

\bibitem{geiger2013vision}
A.~Geiger, P.~Lenz, C.~Stiller, and R.~Urtasun, ``Vision meets robotics: The {KITTI} dataset,'' {\em The International Journal of Robotics Research (IJRR)}, vol.~32, no.~11, pp.~1231--1237, 2013.

\bibitem{lugmayr2022repaint}
A.~Lugmayr, M.~Danelljan, A.~Romero, F.~Yu, R.~Timofte, and L.~Van~Gool, ``{RePaint}: Inpainting using denoising diffusion probabilistic models,'' in {\em Proceedings of the IEEE/CVF Conference on Computer Vision and Pattern Recognition (CVPR)}, pp.~11461--11471, 2022.

\bibitem{emiel2023simple}
E.~Hoogeboom, J.~Heek, and T.~Salimans, ``Simple diffusion: end-to-end diffusion for high resolution images,'' in {\em Proceedings of the International Conference on Machine Learning (ICML)}, pp.~13213--13232, 2023.

\bibitem{ronneberger2015unet}
O.~Ronneberger, P.~Fischer, and T.~Brox, ``{U-Net}: Convolutional networks for biomedical image segmentation,'' in {\em Proceedings of the International Conference on Medical Image Computing and Computer-Assisted Intervention (MICCAI)}, pp.~234--241, 2015.

\bibitem{saxena2023monocular}
S.~Saxena, A.~Kar, M.~Norouzi, and D.~J. Fleet, ``Monocular depth estimation using diffusion models,'' {\em arXiv:2302.14816}, 2023.

\bibitem{xu2021positional}
R.~Xu, X.~Wang, K.~Chen, B.~Zhou, and C.~C. Loy, ``Positional encoding as spatial inductive bias in gans,'' in {\em Proceedings of the IEEE/CVF Conference on Computer Vision and Pattern Recognition (CVPR)}, pp.~13569--13578, 2021.

\bibitem{choi2021toward}
J.~Choi, J.~Lee, Y.~Jeong, and S.~Yoon, ``Toward spatially unbiased generative models,'' in {\em Proceedings of the IEEE/CVF International Conference on Computer Vision (ICCV)}, pp.~14253--14262, 2021.

\bibitem{verbin2022ref}
D.~Verbin, P.~Hedman, B.~Mildenhall, T.~Zickler, J.~T. Barron, and P.~P. Srinivasan, ``{Ref-NeRF}: Structured view-dependent appearance for neural radiance fields,'' in {\em Proceedings of the IEEE/CVF Conference on Computer Vision and Pattern Recognition (CVPR)}, pp.~5481--5490, 2022.

\bibitem{zhang2022curl}
K.~Zhang, Z.~Hong, S.~Xu, and S.~Wang, ``{CURL}: Continuous, ultra-compact representation for {LiDAR},'' in {\em Proceedings of the Robotics: Science and Systems (RSS)}, 2022.

\bibitem{tancik2020fourier}
M.~Tancik, P.~Srinivasan, B.~Mildenhall, S.~Fridovich-Keil, N.~Raghavan, U.~Singhal, R.~Ramamoorthi, J.~Barron, and R.~Ng, ``Fourier features let networks learn high frequency functions in low dimensional domains,'' in {\em Proceedings of the Advances in Neural Information Processing Systems (NeurIPS)}, vol.~33, pp.~7537--7547, 2020.

\bibitem{mildenhall2020nerf}
B.~Mildenhall, P.~P. Srinivasan, M.~Tancik, J.~T. Barron, R.~Ramamoorthi, and R.~Ng, ``{NeRF}: Representing scenes as neural radiance fields for view synthesis,'' in {\em Proceedings of the European Conference on Computer Vision (ECCV)}, pp.~405--421, 2020.

\bibitem{dhariwal2021diffusion}
P.~Dhariwal and A.~Nichol, ``Diffusion models beat gans on image synthesis,'' in {\em Proceedings of the Advances in neural information processing systems (NeurIPS)}, vol.~34, pp.~8780--8794, 2021.

\bibitem{nakashima2018learning}
K.~Nakashima, H.~Jung, Y.~Oto, Y.~Iwashita, R.~Kurazume, and O.~M. Mozos, ``Learning geometric and photometric features from panoramic {LiDAR} scans for outdoor place categorization,'' {\em Advanced Robotics}, vol.~32, no.~14, pp.~750--765, 2018.

\bibitem{schubert2019circular}
S.~Schubert, P.~Neubert, J.~P{\"o}schmann, and P.~Protzel, ``Circular convolutional neural networks for panoramic images and laser data,'' in {\em Proceedings of the IEEE Intelligent Vehicles Symposium (IV)}, pp.~653--660, 2019.

\bibitem{lin2017refinenet}
G.~Lin, A.~Milan, C.~Shen, and I.~Reid, ``Refinenet: Multi-path refinement networks for high-resolution semantic segmentation,'' in {\em Proceedings of the IEEE/CVF Conference on Computer Vision and Pattern Recognition (CVPR)}, pp.~1925--1934, 2017.

\bibitem{triess2020scan}
L.~T. Triess, D.~Peter, C.~B. Rist, and J.~M. Z{\"o}llner, ``Scan-based semantic segmentation of {LiDAR} point clouds: An experimental study,'' in {\em Proceedings of the IEEE Intelligent Vehicles Symposium (IV)}, pp.~1116--1121, 2020.

\bibitem{milioto2019rangenet}
A.~Milioto, I.~Vizzo, J.~Behley, and C.~Stachniss, ``{RangeNet++}: Fast and accurate {LiDAR} semantic segmentation,'' in {\em Proceedings of the IEEE/RSJ International Conference on Intelligent Robots and Systems (IROS)}, pp.~4213--4220, 2019.

\bibitem{behley2019semantickitti}
J.~Behley, M.~Garbade, A.~Milioto, J.~Quenzel, S.~Behnke, C.~Stachniss, and J.~Gall, ``{SemanticKITTI}: A dataset for semantic scene understanding of {LiDAR} sequences,'' in {\em Proceedings of the IEEE/CVF International Conference on Computer Vision (ICCV)}, pp.~9297--9307, 2019.

\bibitem{dowson1982frechet}
D.~C. Dowson and B.~V. Landau, ``The {Fr{\'e}chet} distance between multivariate normal distributions,'' {\em Journal of Multivariate Analysis}, vol.~12, no.~3, pp.~450--455, 1982.

\bibitem{shu20193d}
D.~W. Shu, S.~W. Park, and J.~Kwon, ``{3D} point cloud generative adversarial network based on tree structured graph convolutions,'' in {\em Proceedings of the IEEE/CVF International Conference on Computer Vision (ICCV)}, pp.~3859--3868, 2019.

\bibitem{qi2017pointnet1}
C.~R. Qi, H.~Su, K.~Mo, and L.~J. Guibas, ``{PointNet}: Deep learning on point sets for {3D} classification and segmentation,'' in {\em Proceedings of the IEEE/CVF Conference on Computer Vision and Pattern Recognition (CVPR)}, pp.~652--660, 2017.

\bibitem{chang2015shapenet}
A.~X. Chang, T.~Funkhouser, L.~Guibas, P.~Hanrahan, Q.~Huang, Z.~Li, S.~Savarese, M.~Savva, S.~Song, H.~Su, J.~Xiao, L.~Yi, and F.~Yu, ``{ShapeNet}: An information-rich {3D} model repository,'' {\em arXiv:1512.03012}, 2015.

\bibitem{shan2020simulation}
T.~Shan, J.~Wang, F.~Chen, P.~Szenher, and B.~Englot, ``Simulation-based lidar super-resolution for ground vehicles,'' {\em Robotics and Autonomous Systems (RAS)}, vol.~134, p.~103647, 2020.

\bibitem{kwon2022implicit}
Y.~Kwon, M.~Sung, and S.-E. Yoon, ``Implicit {LiDAR} network: Lidar super-resolution via interpolation weight prediction,'' in {\em Proceedings of the International Conference on Robotics and Automation (ICRA)}, pp.~8424--8430, 2022.

\bibitem{dosovitskiy2017carla}
A.~Dosovitskiy, G.~Ros, F.~Codevilla, A.~Lopez, and V.~Koltun, ``{CARLA}: An open urban driving simulator,'' in {\em Proceedings of the Conference on Robot Learning (CoRL)}, pp.~1--16, 2017.

\end{thebibliography}
